\documentclass[%
 reprint,
 amsmath,amssymb,
 aps,
]{revtex4-2}

\usepackage{graphicx}
\usepackage{dcolumn}
\usepackage{bm}
\usepackage{comment}
\usepackage{color}
\usepackage{colortbl}

\begin{document}

\preprint{APS/123-QED}

\title{
    A Unifying Framework for Information Processing in Stochastically Driven Dynamical Systems
}

\author{Tomoyuki Kubota}
    \email{kubota@ai.u-tokyo.ac.jp}
    \affiliation{Graduate School of Information Science and Technology, The University of Tokyo.}
\author{Hirokazu Takahashi}%
    \affiliation{Graduate School of Information Science and Technology, The University of Tokyo.}%
\author{Kohei Nakajima}%
    \affiliation{Graduate School of Information Science and Technology, The University of Tokyo.}%

\date{\today}

\begin{abstract}
    A dynamical system can be regarded as an information processing apparatus that encodes input streams from the external environment to its state and processes them through state transitions. The information processing capacity (IPC) is an excellent tool that comprehensively evaluates these processed inputs, providing details of unknown information processing in black box systems; however, this measure can be applied to only time-invariant systems. This paper extends the applicable range to time-variant systems and further reveals that the IPC is equivalent to coefficients of polynomial chaos (PC) expansion in more general dynamical systems. To achieve this objective, we tackle three issues. First, we establish a connection between the IPC for time-invariant systems and PC expansion, which is a type of polynomial expansion using orthogonal functions of input history as bases. We prove that the IPC corresponds to the squared norm of the coefficient vector of the basis in the PC expansion. Second, we show that an input following an arbitrary distribution can be used for the IPC, removing previous restrictions to specific input distributions. Third, we extend the conventional orthogonal bases to functions of both time and input history and propose the IPC for time-variant systems. To show the significance of our approach, we demonstrate that our measure can reveal information representations in not only machine learning networks but also a real, cultured neural network. Our generalized measure paves the way for unveiling the information processing capabilities of a wide variety of physical dynamics which has been left behind in nature. 
\end{abstract}

\maketitle


\section{\label{sec:Intro}Introduction}
Dynamical systems driven by external stimuli can be universally found in nature, especially in biology. 
The dynamical aspects of information processing found in biology have long been a source of inspiration for researchers who wish to create a high-speed, energy efficient, and robust real-time information processing device, which resolves a von Neumann bottleneck \cite{stieg2012emergent}. 
Reservoir computing (RC) \cite{jaeger2004harnessing,maass2002real,verstraeten2007experimental} is a bioinspired information processing paradigm that capitalizes on this dynamical perspective and has been widely utilized in various fields in recent years \cite{appeltant2011information,brunner2013parallel,vandoorne2014experimental,du2017reservoir,
torrejon2017neuromorphic,
moon2019temporal,marinella2019efficient,
ludge2019computing
}. 
It consists of a type of learning framework for recurrent neural networks (RNNs), whose intermediate layer is referred to as the reservoir. 
In a reservoir composed of $N$-nodes,
the $i^{\rm th}~(i=1,2,\ldots,N)$ node state $x_{i,t}$ at the $t^{\rm th}$ time step can be written as follows:
\begin{eqnarray}
    x_{i,t+1} &=& f\left(\sum_{j=1}^N w_{ij}x_{j,t} + w_{in,i}u_t\right), 
\end{eqnarray}
where $f$ is the activation function, and $w_{ij}$ and $w_{in,i}$ are the internal and input weights, respectively. 
To emulate the target output $z_t$, we use linear regression to obtain an estimate of $z_t$, $\tilde{z}_t$, as follows:
\begin{eqnarray}
    \tilde{z}_t &=& \tilde{\bm w}_{out}^\top\cdot{\bm x}_t, \label{eq:ztilde}\\
    \tilde{\bm w}_{out} &=& \arg \min_{{\bm w}_{out}}\sum_{t=1}^T \left(z_t-{\bm w}_{out}^\top\cdot{\bm x}_t\right)^2, \label{eq:wtilde}
\end{eqnarray}

\noindent where ${\bm w}_{out}$ and $\tilde{\bm w}_{out}\in\mathbb{R}^N$ are the weight and solution vector for the target output, respectively. 
This learning method leverages the dynamical resource through training without affecting the state of the reservoir but places a constraint on ${\bm x}_t$. 
To perform reproducible computation, RC requires ${\bm x}_t$ to be the same response against identical input time-series (i.e., 
the state needs to be an echo function, which is a function of only the past input time-series ${\bm u}_t=\{u_{t-s}\}_{s=1}^t$). 
This dynamical property is referred to as the echo state property (ESP) \cite{jaeger2002tutorial,yildiz2012re,manjunath2013echo} or the fading memory property (FMP) \cite{maass2002real,maass2004computational,maass2011liquid}, which are slightly different from each other (see the Appendix for further details). 
According to these properties, various activation functions can be used for the reservoir node.
Furthermore, as the reservoir is not limited to a computer-generated system, it can be replaced with a real physical system. 
A reservoir using such a physical system is called a physical reservoir \cite{nakajima2020physical}. 
Some of the above systems have a wide range of dynamics that are not readily found in conventional neural networks.

In the literature, a measure called information processing capacity (IPC) \cite{dambre2012information} has been proposed to quantify the information processing capability of dynamical systems that have ESPs or FMPs. 
The IPC measures the type and quantity of input history that is handled and held in the system by decomposing the system state into an orthogonal basis \cite{martinez2020information,nokkala2020gaussian,akashi2020input}. 
Using the $N$-dimensional state ${\bm x}_t \in \mathbb{R}^N$ and the one-dimensional stochastic input $\zeta_{t}\in\mathbb{R}$ at the $t^{\rm th}~(t\in\mathbb{Z})$ time step, 
the input-driven dynamical system (IDS) determines the next state, as shown below: 
\begin{eqnarray}
    {\bm x}_{t+1} = {\bm f}\left( {\bm x}_{t}, \zeta_{t} \right), \label{eq:x}
\end{eqnarray}
where ${\bm f}$ maps $\mathbb{R}^{N}\times\mathbb{R} \rightarrow \mathbb{R}^N$. 
The IPC evaluates the emulation ability of the $i^{\rm th}$ target output $z^{(i)}_t\in\mathbb{R}~(i=1,2,\ldots)$ from ${\bm x}_t~(t=1,2,\ldots,T)$. 
$z^{(i)}_t~(i=1,2,\ldots)$ is represented by the product of the $n_s^{(i)}$-th order polynomial of the random variable delayed by $s$ steps, $\zeta_{t-s}$, 
\begin{eqnarray}
    z^{(i)}_t &=& \prod_{s=1}^\infty \mathcal{F}_{n_s^{(i)}} (\zeta_{t-s}), \label{eq:z}
\end{eqnarray}
where $\mathcal{F}_{n}(\zeta)$ represents the $n^{\rm th}$ order orthogonal polynomial of $\zeta$. 
From Eqs. (\ref{eq:ztilde}) and (\ref{eq:wtilde}), we obtain an estimate of $z^{(i)}_t$, $\tilde{z}^{(i)}_t$. 
When $z_t$ is an orthogonal function of the independent variables $\{\zeta_{t-s}\}_{s=1}^\infty$, 
the IPC is defined using a normalized emulation error of the reservoir, as follows: 
\begin{eqnarray}
    C\left( {\bm X}, {\bm z}^{(i)} \right) 
    &=& 1 - \frac{\min_{\bm w} \sum_{t=1}^T \left( z^{(i)}_t - {\bm w}^\top {\bm x}_{t} \right)^2}{ \sum_{t=1}^T \left( z^{(i)}_t \right)^2 } \nonumber \\
    &=& \frac{{\bm z}^{(i)\top}{\bm X}\left( {\bm X}^\top{\bm X}\right)^{-1}{\bm X}^\top{\bm z}^{(i)}}{{\bm z}^{(i)\top}{\bm z}^{(i)}}, \label{eq:IPC1}
\end{eqnarray}
where ${\bm w}\in\mathbb{R}^N$, ${\bm X} = \left[ {\bm x}_1 \cdots {\bm x}_T \right]^\top \in \mathbb{R}^{T \times N}$, and ${\bm z}^{(i)} = \left[ z^{(i)}_1 \cdots z^{(i)}_T \right]^\top \in \mathbb{R}^{T}$ are the weight vector, state, and target output, respectively. 
In this case, the uniform random variable and the Legendre polynomial are the stochastic variable $\zeta$ and orthogonal polynomial $\mathcal{F}_n(\zeta)$, respectively, 
although the combination is not restricted. 
For example, a Gaussian random variable and a Hermite polynomial are also suitable \cite{dambre2012information}. 
Therefore, the IPC is a measure used to evaluate the input information held by the state with the emulation ability of the orthogonal basis.

In this connection, there is a theory about a deterministic dynamical system with a stochastic input $\zeta_{t}$ in a different context. 
The system can be described as an operator of $\{\zeta_{t-s}\}_{s=1}^\infty$ according to the polynomial chaos expansion \cite{wiener1938homogeneous}, which is a series expansion using the target outputs of IPC as the bases (i.e., multivariate orthogonal polynomials of the random variables, $z_t^{(i)}$, described in Eq. [\ref{eq:z}]). 
The polynomial chaos expansion has been frequently utilized to determine the evolution of uncertainty in a dynamical system when there is probabilistic uncertainty in the system parameters \cite{xiu2002wiener,oladyshkin2012data}. 
These multivariate polynomials are referred to as polynomial chaoses (PCs), and the space spanned by PCs is called homogeneous chaos \cite{wiener1938homogeneous}, expressed as 
\begin{eqnarray}
    {\bm x}_t = \sum_{i=1}^{\infty} {\bm c}_i z_t^{(i)}, \label{eq:expanded_x}
\end{eqnarray}
where ${\bm c}_i\in\mathbb{R}^N$ is the $i^{\rm th}$ coefficient vector. 
If the input $\zeta_t$ follows a certain distribution, the PC is determined based on its orthogonality. 
Let ${\bm\zeta}$ be a vector notation of $\zeta$ sampled $T$ times ${\bm\zeta}=\{\zeta_1,\ldots,\zeta_T\}$. 
If a weighting function $w({\bm\zeta})$ specific to the PCs $\{z_t^{(i)}\}_{i=1}^\infty$ exists, the following orthogonality relations should be satisfied: 
\begin{eqnarray}
    \left< z_t^{(i)} z_t^{(j)} \right> &=& \left<(z_t^{(i)})^2\right>\delta_{ij}, \label{eq:inner_product_z} \\
    \left< f({\bm\zeta})g({\bm\zeta}) \right> &=& \sum_{\bm\zeta} w({\bm\zeta})f({\bm\zeta})g({\bm\zeta}), \label{eq:inner_product_define}
\end{eqnarray}
where $\delta_{ij}$ is the Kronecker delta function. 
The PC expansion can use various combinations of input and polynomials. 
The generalized polynomial chaos (gPC)\cite{xiu2002wiener} supplies the PC for specific combinations---e.g., the Hermite polynomial for Gaussian distributions, the Legendre polynomial for uniform distributions, and the Charlier polynomial for Poisson distributions are available as $\mathcal{F}_n(\zeta)$. 
Furthermore, the arbitrary polynomial chaos (aPC) \cite{oladyshkin2012data} is the PC for random variables following an arbitrary probability distribution by using the Gram-Schmidt orthogonalization procedure (See the Appendix for both schemes).

As previously described, the IPC and PC expansion have a number of similarities and differences. 
Both schemes use the IDS with stochastic inputs and multivariate orthogonal polynomials, whereas the types of input distributions and orthogonal polynomials are not as limited in the IPC as they are in the PC. 
One objective of this paper is to establish a connection between the IPC and the IDS expanded with PCs. 
Thus, we first aim to reveal this relationship by deriving the IPC from the state expanded with PCs. 
Second, to enlarge the applicable range of IPC, we extend this relationship for time-variant systems. 
So far, the IPC assumed that the system is a function of only the input time-series. 
However, a solution of the dynamical system in Eq. (\ref{eq:x}) is a function of the input time-series and time with a given initial state. 
Introducing time-dependent orthogonal bases to PCs, we aim to derive the IPC for time-variant systems and illustrate that input information processing is performed by coupling the terms of time and input time-series. 
Finally, to demonstrate the potential of our approach, we apply our theory to three cases---a model that is frequently used as a benchmark task in the context of temporal machine learning, an artificial neural network, and a real, cultured neural network---to reveal their information processing within the systems.

\section{Methods}
\subsection{Singular value decomposition}
Singular value decomposition (SVD) breaks down state $\bm X\in\mathbb{R}^{T\times N}$ into
\begin{eqnarray}\label{eq:svd}
    {\bm X}={\bm P}{\bm \Sigma}{\bm Q}^\top, 
\end{eqnarray}
where ${\bm P} \in \mathbb{R}^{T\times r}$ and ${\bm Q} \in \mathbb{R}^{N\times r}$ are matrices whose column and row vectors, respectively, are singular vectors, ${\bm \Sigma}={\rm diag}\{\sigma_1,\ldots,\sigma_r\} \in \mathbb{R}^{r\times r}$ is a diagonal matrix containing the singular values $\sigma_i~(i=1,\ldots,r)$, and $r(\le N<T)$ is the rank of ${\bm X}$.

\subsection{ESN}
Let the $i^{\rm th}~(i=1,2,\ldots,N)$ state of the ESN at the $t^{\rm th}$ step be $x_{i,t}$. 
The state equation is given as follows: 
\begin{eqnarray}
    x_{i,t+1} &=& f\left(\rho \sum_{j=1}^N w_{i,j} x_{j,t} + \iota \omega_{i} \zeta_{t} \right), \label{eq:esn} 
\end{eqnarray}
where $w_{ij}$ was initialized with the uniform random number in the range of $[-1,1]$ and multiplied by a constant so that the maximum eigenvalue of the matrix ${\bm W}=[w_{ij}]$ was 1. 
The input weight $\omega_{i}$ was also set to a uniform random number in the range of $[-1,1]$, and $\rho$, $\iota(=0.1)$, and $N(=50)$ represent the spectral radius of $\rho {\bm W}$, input intensity, and the number of nodes, respectively. 
$f$ denotes the activation function, three types of which were used to solve the NARMA10 task \cite{jaeger2003adaptive,verstraeten2007experimental}: a linear function 
\begin{eqnarray}
    f(y) = y, \label{eqA:esn_linear}
\end{eqnarray}
a hyperbolic tangent function
\begin{eqnarray}
    f(y) = \tanh(y), \label{eqA:esn_tanh}
\end{eqnarray}
and an analog integrator function
\begin{eqnarray}
    f(y) = \left(1-\frac{1}{\tau}\right)x_{i,t} + \frac{1}{\tau} \tanh(y), \label{eqA:esn_integrator}
\end{eqnarray}
where $\tau$ was set to 1.25.

\subsection{One-dimensional ESN}
The one-dimensional ESN is described as 
\begin{eqnarray}
    x_{t+1} &=& \tanh(\rho x_t+u_t), \label{eq:1dim_esn_x} \\
    u_t &=& \mu+\sigma\zeta_t, \label{eq:1dim_esn_u}
\end{eqnarray}
where $x_t$, $u_t$, and $\zeta_t$ are the state, input, and random variable at the $t^{\rm th}$ step, respectively, and $\rho$ and $\sigma$ are an inner weight and input intensity, respectively. 
To treat the bounded and time-invariant state, we chose $\rho=0.95$ and $\mu=0$.

\subsection{NARMA10 benchmark task}
Non-linear autoregressive moving average (NARMA) tasks were introduced to test the performance of RNNs \cite{atiya2000new}. 
Within these tasks, the target output $y_t$ is generated with the NARMA model $y_{t+1}=g(y_t,y_{t-1},\ldots,\zeta_t,\zeta_{t-1},\ldots)$ and input series $\{\zeta_t\}_{t=1}^T$. 
The RNN receives the same input $\zeta_t$ and emulates $y_t$ by modifying its weights.
NARMA tasks, including NARMA2, NARMA10 \cite{atiya2000new}, and NARMA30 \cite{verstraeten2009quantification}, evaluate the ability to emulate the model. 
In particular, NARMA10 has been widely used as a benchmark task for RC \cite{
jaeger2003adaptive,rodan2010minimum,verstraeten2007experimental,
nakajima2019boosting,
appeltant2011information,
paquot2012optoelectronic,
dale2016evolving,
nakajima2013computing,nakajima2018exploiting,
burger2015hierarchical,fujii2017harnessing,hermans2016embodiment,yin2012developmental,okumura2019experimental,barazani2020microfabricated,
bianchi2017multiplex,
duport2016fully,
hermans2012recurrent,
inubushi2017reservoir,
kan2021simple}. 
Some specific examples include testing the performance of ESNs \cite{jaeger2003adaptive,rodan2010minimum,verstraeten2007experimental} as well as evaluating the computational capability of physical systems, such as quantum systems \cite{nakajima2019boosting,fujii2017harnessing,tran2020higher}, analog circuits \cite{appeltant2011information}, opto-electronic architectures \cite{paquot2012optoelectronic}, carbon nanotubes \cite{dale2016evolving}, soft robotic systems \cite{nakajima2013computing,nakajima2018exploiting,torres2019information}, and other dynamical systems \cite{burger2015hierarchical,fujii2017harnessing,hermans2016embodiment,yin2012developmental,barazani2020microfabricated,okumura2019experimental}. 
Thus, NARMA10 is one of the most representative benchmark tasks and is broadly utilized to compare and evaluate the computational capabilities of dynamical systems.

Let the state and input at the $t^{\rm th}$ step be $y_t$ and $u_t$, respectively. 
The NARMA10 model is given by 
\begin{eqnarray}
    y_{t+1} &=& \alpha y_t + \beta y_t \sum_{s=0}^{9}y_{t-s} + \gamma u_{t} u_{t-9} + \delta, \label{eq:narma10} \\
    u_t &=& \mu + \kappa \zeta_t, \label{eq:narma10_ut}
\end{eqnarray}
where the default constant parameters $(\alpha,\beta,\gamma,\delta)$ are set to $(0.3,\ 0.05,\ 1.5,\ 0.1)$, $\zeta_t$ is the random variable at the $t^{\rm th}$ step and follows a uniform distribution in the $[-1,1]$ interval, while $\mu$ and $\kappa$ are the average of $\zeta_t$ and the input intensity parameter, respectively. 
This paper uses two ranges: $u_t\in[0,\sigma]$ ($\mu=\kappa=\sigma/2$) and $u_t\in[-\sigma,\sigma]$ ($\mu=0$, $\kappa=\sigma$). 
All the initial values of $y_t~(t=0,1,\ldots,9)$ were set to zero, except when the basin of attraction was examined.

\subsection{The limit cycle system}
The simple limit cycle system with radius $r$ and azimuth $\theta$ in polar coordinates \cite{strogatz2001nonlinear} is discretized as follows: 
\begin{eqnarray}
    r_{t+1} &=& (1+\tau)r_t - \tau r_t^3 + \tau u_t, \label{eq:limit_cycle_r}\\
    \theta_{t+1} &=& \theta_t + \tau \omega, \label{eq:limit_cycle_theta}\\
    u_t &=& \mu + \sigma\zeta_t, \label{eq:limit_cycle_u}
\end{eqnarray}
where $\omega=2\pi/3$ and $\tau=0.1$ are the angular velocity and time step width, respectively, and $\zeta_t$ is the uniform random number in the $[-1,1]$ interval. 
Therefore, the input $u_t$ follows the uniform distribution in the $[\mu-\sigma,\mu+\sigma]$ range ($\mu=0.2$ and $\sigma=1.5$) and is applied in the radial direction. 
The Cartesian coordinates are given by $x_t=r_t\cos\theta_t$ and $y_t=r_t\sin\theta_t$.

\section{\label{sec:result}Results}
\subsection{The equivalence of the IPC and coefficient in PC expansion}
To show the relationship between the IPC and PC expansion, we derive the IPC from the state expanded in terms of PCs. 
First, we transform the IPC into a simpler form using SVD, which reduces the $N$ state time-series to $r(\le N)$ linearly independent time-series vectors ${\bm p}_j\in\mathbb{R}^T~(j=1,\ldots,r)$. 
Using the decomposed state, we can rewrite the IPC relative to ${\bm z}^{(i)}$ as 
\begin{eqnarray}\label{eq:IPC2}
    C\left( {\bm X}, {\bm z}^{(i)} \right) = \sum_{j=1}^r \left( {\bm p}_j^\top {\bm \phi}^{(i)} \right)^2,
\end{eqnarray}
where ${\bm \phi}^{(i)}={\bm z}^{(i)}/||{\bm z}^{(i)}||$ is the normalized output. 
Next, assuming that ${\bm x}_t$ can be expanded with PCs, the state ${\bm X}$ is described as 
\begin{eqnarray} \label{eq:PhiC}
    {\bm X} = \sum_{i=1}^{\infty} {\bm \phi}^{(i)} \cdot \hat{\bm c}_i^\top
    = {\bm\Phi}\hat{\bm C}^\top, 
\end{eqnarray}
where ${\bm \Phi} = \left[ {\bm \phi}^{(1)}~{\bm \phi}^{(2)} \cdots \right]$ and $\hat{\bm C} = [\hat{\bm c}_1~\hat{\bm c}_2\cdots]^\top$ are the basis matrix and coefficient matrix, respectively. 
Comparing the decomposed state ${\bm X} = {\bm P}{\bm\Sigma}{\bm Q}^\top$ (see Methods) and Eq. (\ref{eq:PhiC}), we obtain the matrix
\begin{eqnarray}
    {\bm P} &=& {\bm\Phi}{\bm\Lambda}^\top, \label{eq:P}
\end{eqnarray}
where ${\bm P} = [{\bm p}_1~{\bm p}_2\cdots{\bm p}_r]^\top$ and ${\bm\Lambda}={\bm \Sigma}^{-1}{\bm Q}^\top\hat{\bm C}=[{\bm\lambda}_1~{\bm\lambda}_2\cdots]$ are a matrix form of linearly independent vectors and a constant matrix, respectively, while ${\bm\lambda}_i\in\mathbb{R}^r$. 
From Eqs. (\ref{eq:IPC2}) and (\ref{eq:P}), the IPC becomes
\begin{eqnarray} \label{eq:IPC_lambda}
    C({\bm X},{\bm z}^{(i)}) = ||{\bm \lambda}_i||^2. 
\end{eqnarray}
Because ${\bm P}$ is also written as ${\bm P} = \sum_{i=1}^\infty {\bm\phi}^{(i)}\cdot{\bm\lambda}_i^\top$, 
Eq. (\ref{eq:IPC_lambda}) illustrates that the computation of the $i^{\rm th}$ IPC is equivalent to expanding the state with the PCs and calculating the squared norm of the coefficient of the $i^{\rm th}$ PC of the temporal bases ${\bm p}_j$ expanded by the PCs.

In addition, the IPC has an important property of summation. 
The total capacity is described as 
\begin{eqnarray}\label{eq:Ctot}
    C_{tot} = \sum_{i=1}^\infty C\left( {\bm X}, {\bm z}^{(i)} \right) 
    = \sum_{j=1}^r || {\bm \Phi}^\top {\bm p}_j ||^2. 
\end{eqnarray}
Eq. (\ref{eq:Ctot}) yields the sum of the squared norm of ${\bm p}_j~(j=1,\ldots,r)$ projected into an infinite-dimensional space that contains the orthogonal vectors ${\bm \phi}^{(i)}~(i=1,2,\ldots)$; 
therefore, if all of the ${\bm p}_j$ are functions of only past input history, $||{\bm\Phi}^\top{\bm p}_j||$ becomes one, resulting in 
\begin{eqnarray}\label{eq:maxCtot}
    C_{tot}=r ~ (\le N). 
\end{eqnarray}
We call this condition an {\it integrity property}. 
Under the assumption that the state is a function of only past input history, IPCs hold the integrity property in information processing. 
These results provide a new perspective that the coefficient of PC expansion represents the amount of processed input.

\subsection{Demonstration of comprehensive computational capabilities using general input distribution}
\begin{figure*}[tb]
    \centering
    \includegraphics[scale=0.85]{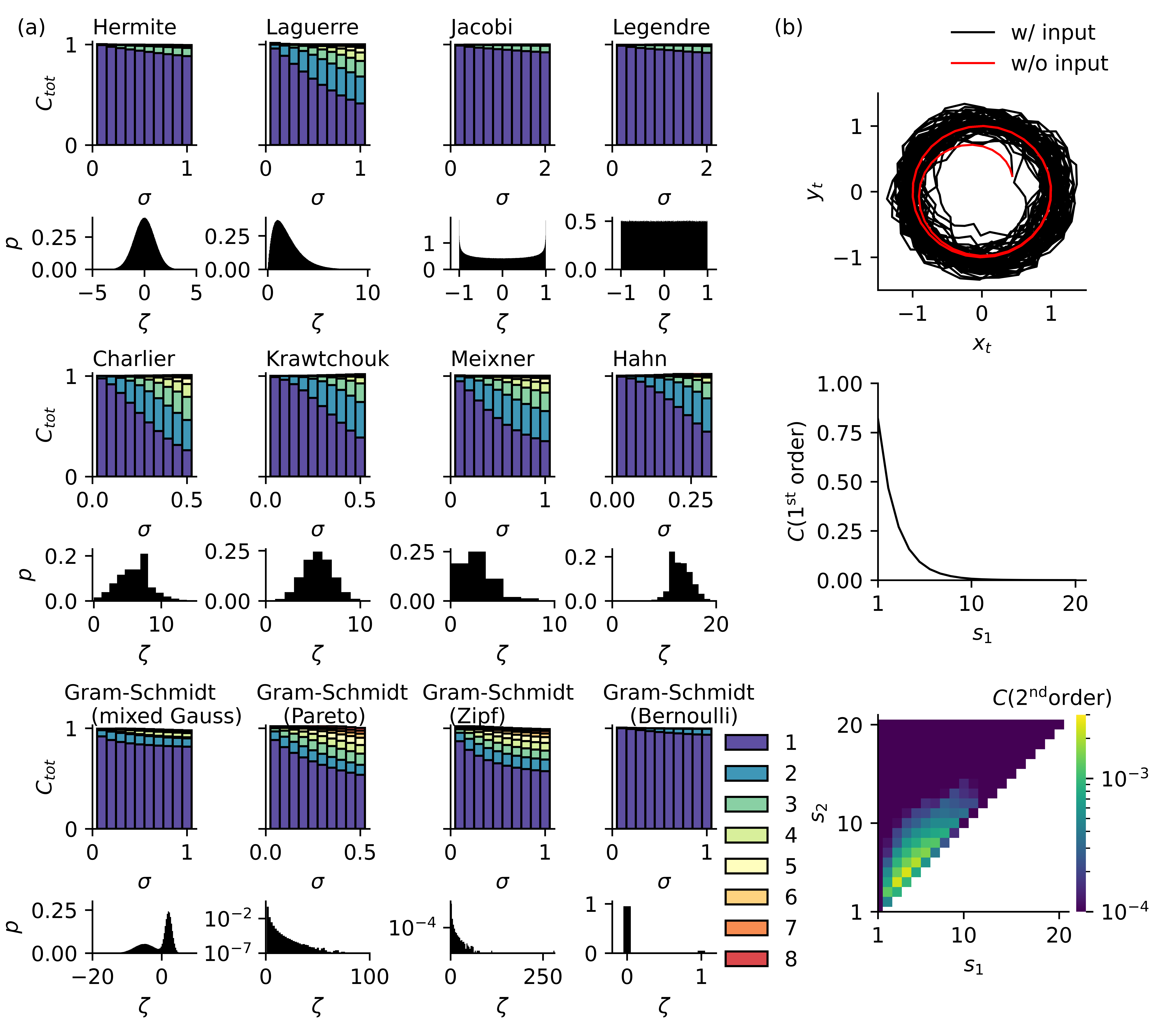}
    \caption{
        Demonstration of theories. 
        (a) IPC breakdowns of the one-dimensional ESN with eight types of target generalized/arbitrary PCs of a random variable $\zeta_t$: 
        Hermite-chaos with Gaussian random variables, 
        Laguerre-chaos with gamma random variables, 
        Jacobi-chaos with beta random variables, 
        Legendre-chaos with uniform random variables, 
        Charlier-chaos with Poisson random variables, 
        Krawtchouk-chaos with binomial random variables, 
        Meixner-chaos with negative binomial random variables, 
        Hahn-chaos with hypergeometric random variables, 
        and Gram-Schmidt-chaos with 
        mixed-Gauss random variables, 
        Pareto random variables, 
        Zipf random variables, and 
        Bernoulli random variables. 
        The $n^{\rm th}$-order capacities $(n=1,\ldots,8)$ in the total capacity $C_{tot}$ are summarized. 
        The variable $p$ represents the probability distribution of $\zeta_t$. 
        (b) The TIPC of the limit cycle system in Eqs. (\ref{eq:limit_cycle_r})--(\ref{eq:limit_cycle_u}). 
        The phase plane of $x_t$ and $y_t$ in the system (upper panel). 
        The systems without input ($\mu=\sigma=0$; red line) and with input ($\mu=0.2,\sigma=1.5$; black line) are shown. 
        The first-order capacities of $P_1(\zeta_{t-s_1})\cos(\omega\tau t+\alpha_{1,s_1})$ with the delay step $s_1$ (middle panel). 
        The second-order capacities relative to delay steps $s_1$ and $s_2$ (lower panel). 
        The diagonal ($s_1=s_2$) and the upper left triangle ($s_1<s_2$) dots show the capacities of $P_2(\zeta_{t-s_1})\cos(\omega\tau t+\alpha_{2,s_1})$ and $P_1(\zeta_{t-s_1})P_1(\zeta_{t-s_2})\cos(\omega\tau t+\alpha_{2,s_1,s_2})$, respectively. 
        In the white area, the TIPC calculation is omitted. 
    }
    \label{fig:theory_demo}
\end{figure*}

To illustrate that various orthogonal polynomials can be used as target outputs for the IPC, 
we examine the total capacities for eight target gPCs and four target aPCs using one-dimensional ESNs, which is a time-invariant system. 
FIG. \ref{fig:theory_demo}(a) shows the IPC breakdowns with eight types of gPCs in the Askey scheme. 
The utilized gPCs of random distributions are the Hermite-chaos of a Gaussian distribution, Laguerre-chaos of a gamma distribution, Jacobi-chaos of a beta distribution, Legendre-chaos of a uniform distribution, Charlier-chaos of a Poisson distribution, Krawtchouk-chaos of a binomial distribution, Meixner-chaos of a negative binomial distribution, and Hahn-chaos of a hypergeometric distribution. 
The total IPCs are all one, suggesting that the gPCs form a complete orthogonal system with any combination of distributions, and orthogonal polynomials and can be used for the IPC.

Furthermore, to demonstrate that aPCs can also be used as the target outputs for the IPC, we estimated the IPCs with the ESN given four types of inputs following a mixed Gaussian distribution, Pareto distribution, Zipf distribution, and Bernoulli distribution, which do not follow the Askey scheme. 
To investigate the IPC of the ESN, we built Gram-Schmidt-chaoses. 
As shown in FIG. \ref{fig:theory_demo}(a), the total IPCs were one, indicating that the aPCs formed a complete orthogonal system with the input distribution and Gram-Schmidt-chaoses and can be used for the IPC.

These results suggest that both the gPC and aPC are suitable as the target output of the IPC.

\subsection{Extending the integrity in information processing to a time-variant domain}
The information processing integrity shown so far holds only if the system is a function of a past input series. 
Using the derived relationship between the IPC and PC expansion, we extend the application range of IPC to time-varying IDSs. 
First, we introduce a classification of IDSs based on the conditions imposed on the solution of Eq. (\ref{eq:x}). 
From the connection between the IPC and PC, the IPC is calculated by extracting the second-order process ${\bm x}_t$ in the $1\le t\le T$ range from the original sequence, 
which is obtained according to Eq. (\ref{eq:x}). 
By recursively using Eq. (\ref{eq:x}) from $t=0$, the solution is clearly determined from the time $t$, input sequence ${\bm \zeta}_t = \{\zeta_{t-s}\}_{s=1}^{t}$, and initial state ${\bm x}_0$, as follows: 
\begin{eqnarray}
    {\bm x}_t = {\bm \xi}(t,{\bm \zeta}_t; {\bm x}_0), \label{eq:x_solution}
\end{eqnarray}
where ${\bm \xi}$ is determined by $\bm f$ in Eq. (\ref{eq:x}). 
However, the state is described as an operator of only $\{\zeta_{t-s}\}$, showing that the PC expansion assumes that ${\bm x}_t$ is time-invariant. 
Therefore, two conditions are imposed on the time-series. 
First, every $i^{\rm th}~(i=1,\ldots,N)$ state time-series $x_{i,t}~(t=1,\ldots,T)$ must be a second-order process. 
The specific condition is that the second moment of $x_{i,t}$ must be finite:
\begin{eqnarray}
    \left<x_{i,t}^2\right> < \infty~(i=1,\ldots,N), 
\end{eqnarray}
where $x_{i,t}$ is the $i^{\rm th}$ element of ${\bm x}_t$, $\left<x_{i,t}^2\right>=\sum_{t=1}^T(x_{i,t}-\bar{x}_i)^2/T$, and $\bar{x}_i = \sum_{t=1}^T x_{i,t}/T$. 
Finiteness is a prerequisite for expansion convergence in the sense of $L_2$ due to the Cameron-Martin theorem. 
Second, the solution ${\bm\xi}$ should be time-invariant. 
The extracted time-series needs to be described only with the input time-series ${\bm\zeta}_t$. 
Thus, ${\bm x}_t={\bm\xi}(t,{\bm\zeta}_t;{\bm x}_0)$ is time-invariant if 
\begin{eqnarray}
    {\bm\xi}(t-\tau,{\bm\zeta}_{t-\tau};{\bm x}_0) &=& {\bm\xi}(t,{\bm\zeta}_{t-\tau};{\bm x}_0) ~ {\rm for}~\tau\in\mathbb{Z}. 
\end{eqnarray}
Therefore, the PC expansion assumes the non-divergence and time-invariance of the system.

In contrast, if ${\bm x}_t$ is time-variant, the homogeneous chaos is no longer a complete orthogonal system, and ${\bm x}_t$ is represented by the time-dependent polynomial chaos (TDPC) as 
\begin{eqnarray} \label{eq:TDHC}
    {\bm x}_t = \sum_{i=1}^\infty {\bm \gamma}_{i}({\bm x}_0) \psi_{t}^{(i)}\phi_t^{(i)}, 
\end{eqnarray}
where ${\bm \gamma}_{i}({\bm x}_0)\in\mathbb{R}^N$, $\phi_{t}^{(i)}$, and $\psi_t^{(i)}$ are the $i^{\rm th}$ coefficient vector depending on ${\bm x}_0$, polynomial chaos, and the time-dependent basis, respectively. 
We can construct a complete orthogonal system from $t$ and $\{\zeta_{t-s}\}_{s=1}^\infty$ by adding bases of $t$ that are orthogonal to the PCs of $\{\zeta_{t-s}\}$. 
In the following discussion, we assume that ${\bm x}_0$ is given and fixed. 
By converting the summation over the input time-series ${\bm\zeta}_t$ into a summation over time $t$, Eq. (\ref{eq:inner_product_z}) can be rewritten as follows:
\begin{eqnarray}
    \left<f_t^{(i)}f_t^{(j)}\right> &=& \left<(f_t^{(i)})^2\right>\delta_{ij}, \label{eq:inner_product_re1} \\
    \left<f_t^{(i)}f_t^{(j)}\right> &=& \sum_{t=1}^T f_t^{(i)}f_t^{(j)}, \label{eq:inner_product_re2} 
\end{eqnarray}
and can be satisfied by the PCs. 
Next, we define the time-dependent basis that satisfies this inner product. 
For example, $\{1,\cos\omega t,\sin\omega t,\cos 2\omega t,\sin 2\omega t,\ldots\}$ ($\omega=2\pi/T$) is used for the Fourier series expansion and is a complete orthogonal basis of time. 
Since $\phi_t^{(i)}$ and $\psi_t^{(j)}$ are uncorrelated with each other, we expect that the following orthogonality relation will be satisfied for a sufficiently long period $T$: 
\begin{eqnarray}
    \left<\phi_t^{(i)}\psi_t^{(i)}\phi_t^{(j)}\psi_t^{(j)}\right> &=& \left<\left(\phi_t^{(i)}\psi_t^{(i)}\right)^2\right>\delta_{ij}. 
\end{eqnarray}
In this paper, we call the time and input time-series-dependent basis in Eq. (\ref{eq:TDHC}), $\psi_t^{(i)}\phi_t^{(i)}$, the TDPC and define the space spanned by the time-dependent polynomial chaoses as time-dependent homogeneous chaos (TDHC).

The IPCs of ${\bm x}_t$ in the TDHC can be estimated by replacing the target output ${\phi}_t^{(i)}$ in Eq. (\ref{eq:IPC2}) with the TDPC $\psi_t^{(i)}\phi_t^{(i)}$. 
We define the IPC with the target TDPCs as temporal information processing capacity (TIPC). 
As with the IPC of a time-invariant system, the TIPC of a time-variant system is equivalent to expanding the state with TDPCs and calculating the squared norm of the coefficient of the $i^{\rm th}$ TDPC of the temporal bases ${\bm p}_j~(j=1,\ldots,r)$ expanded by the TDPCs. 
Note that to remove the terms that do not include the input, the time-average of ${\bm x}_t$ and the time-varying terms---for example, $A_n\cos(2\pi f_nt+\theta_n)~(n=1,2,\ldots)$, where the amplitude $A_n$, the frequency $f_n$, and the phase $\theta_n$ are estimated by the Fourier transform---are subtracted. 
Since TDPCs constitute a complete orthogonal system, the norm of each of the $r$-time-series vectors in the space is one, and the total TIPC retains $r$. 
This extension can reveal aspects of the information processing performed by time-variant systems.

\subsection{Demonstration of complete computational capabilities in a time-variant system}
To demonstrate the extension of the IPC, we show the TIPC of a two-dimensional limit cycle system. 
As shown in FIG. \ref{fig:theory_demo}(b), input that follows a uniform distribution forces the system to fluctuate around the non-input state. 
Using this time-variant system, we calculated the TIPCs, whose TDPCs were constructed using the PC $\{\phi_t\}$ for $n_s~(\sum_s n_s<5)$, $s<10$, and temporal basis $\psi_t\in\{\cos\Omega t,\sin\Omega t,\cos 2\Omega t,\sin 2\Omega t,\ldots,\cos \frac{T}{2} \Omega t,\sin \frac{T}{2} \Omega t\}$, where $\Omega=2\pi/T$. 
From the estimated TIPCs, the state on the two-dimensional plane ${\bm x}_t = [x_t ~ y_t]^\top$ can be expressed as follows: 
\begin{eqnarray}
    {\bm x}_t &=& \sum_{s_1=1}^{18} {\bm p}_{s_1} P_1(\zeta_{t-s_1}) \cos\left(\omega\tau t + \alpha_{1,s_1}\right) \nonumber\\ 
    &+& \sum_{s_1,s_2} {\bm q}_{s_1,s_2} P_1(\zeta_{t-s_1}) P_1(\zeta_{t-s_2}) \cos\left(\omega\tau t + \alpha_{2,s_1,s_2}\right) \nonumber\\
    &+& \sum_{s_1=2}^{8} {\bm q}_{s_1} P_2(\zeta_{t-s_1}) \cos\left(\omega\tau t + \alpha_{2,s_1}\right), \label{eq:limit_cycle_tdpc}
\end{eqnarray}
where $P_n(\zeta)$ represents the $n^{\rm th}$-order Legendre polynomial, 
and the phases $\alpha_{1}$ and $\alpha_{2}$ depend on the initial values. 
Note that although TIPCs depend on the initial values of the system in general, the final outcome of TIPCs is the same in this case because the choice of the initial values does not affect the coefficient vectors of Eq. (\ref{eq:limit_cycle_tdpc}). 
The coefficient vectors for the first and second-order terms are ${\bm p}$ and ${\bm q}\in\mathbb{R}^2$, respectively, indicating that the TIPCs of the system are composed of capacities of the product of the time-varying basis vectors $\psi_t\in\{\cos\omega\tau t,~\sin\omega\tau t\}$ and PCs $\{P_n(\zeta_{t-s})\}$.

In addition, in the case of using the conventional IPC, the total IPC saturates at the rank of the state matrix, $r$, only in the system with the negative maximum Lyapunov exponent \cite{dambre2012information}. 
In the limit cycle system we adopted, the Lyapunov exponents of Eqs. (\ref{eq:limit_cycle_r})--(\ref{eq:limit_cycle_u}) are zero in the azimuthal and negative in the radial direction; 
therefore, the maximum Lyapunov exponent is zero. 
Furthermore, the system does not satisfy the ESP because it is time-variant, and the phases $\alpha_1$ and $\alpha_2$ depend on the initial values. 
Although the system, whose rank is two, does not satisfy these conventional conditions, the total TIPC saturates at $C_{tot}=1.987$ ($1^{\rm st}$-order, $C_{tot}=1.952$; $2^{\rm nd}$-order, $C_{tot}=3.45\times10^{-2}$) as shown in FIG. \ref{fig:theory_demo}(b).

Therefore, these results suggest that the information processing that was lacking with conventional IPC can be measured by adding time-varying bases, and the total TIPC can reach the rank even if the maximum Lyapunov exponent is not negative or the system state depends on the initial values.

\subsection{Application \#1: the benchmark task}
To illustrate the usefulness of our theory, we demonstrate the information processing capabilities of three systems. 
First, we analyze the computational capabilities required to emulate a simple model for a time-series benchmark test, which is a well-known NARMA10 model. 
The NARMA10 task is widely utilized to evaluate the computational capabilities of dynamical systems, but the meaning of predicting this model is unknown. 
We classified the model with certain parameter regions to be time-invariant (see the Appendix). 
In FIG. \ref{fig:ipc_narma10}, $p$ expresses the probability of not diverging as a function of $\sigma$ for different random series $\{\zeta_t\}$, and the model is stable with certain parameters. 
Using the non-divergent model, we estimated its IPCs for the target Legendre-chaoses with delayed time step $s(<16)$ and degree $n_s~\left(\sum_s n_s<9\right)$.

Since the NARMA10 model is a one-dimensional system, the total capacity is one, but the breakdown of the IPC changes with some parameters. 
FIG. \ref{fig:ipc_narma10}(a) and (b) show that the capacities using the uniform random variable in an asymmetric range $u_t\in[0,\sigma]$ differ significantly from those using the input in a symmetric range $u_t\in[-\sigma,\sigma]$. 
The capacity breakdown with the symmetric input includes only the second-order capacities ($\sum_s n_s=2$). 
In contrast, the capacities with the asymmetric input contain the first-order ones because the input term emerges as $u_{t-9}u_t = \sigma^2/4 (P_1(\zeta_{t})P_1(\zeta_{t-9})+P_1(\zeta_t)+P_1(\zeta_{t-9})+1)$ in Eq. (\ref{eq:narma10}), including the first-order terms of $\zeta$. 
Hence, in the case of using an asymmetric input, we can regard the model as the system receiving the first-order inputs, which are retained for a few steps. 
These results suggest that the input should be changed according to the dynamical system when one uses the NARMA10 task; 
for example, as the nodes of an ESN are represented by an odd function, such as a hyperbolic tangent, the ESN has only odd capacities \cite{dambre2012information}. 
From this property and our results, the ESN with $u_t\in[0,\sigma]$ emulates the NARMA10 model, whereas the one with $u_t\in[-\sigma,\sigma]$ does not predict it at all.

\begin{figure}[t]
    \centering
    \includegraphics[scale=0.85]{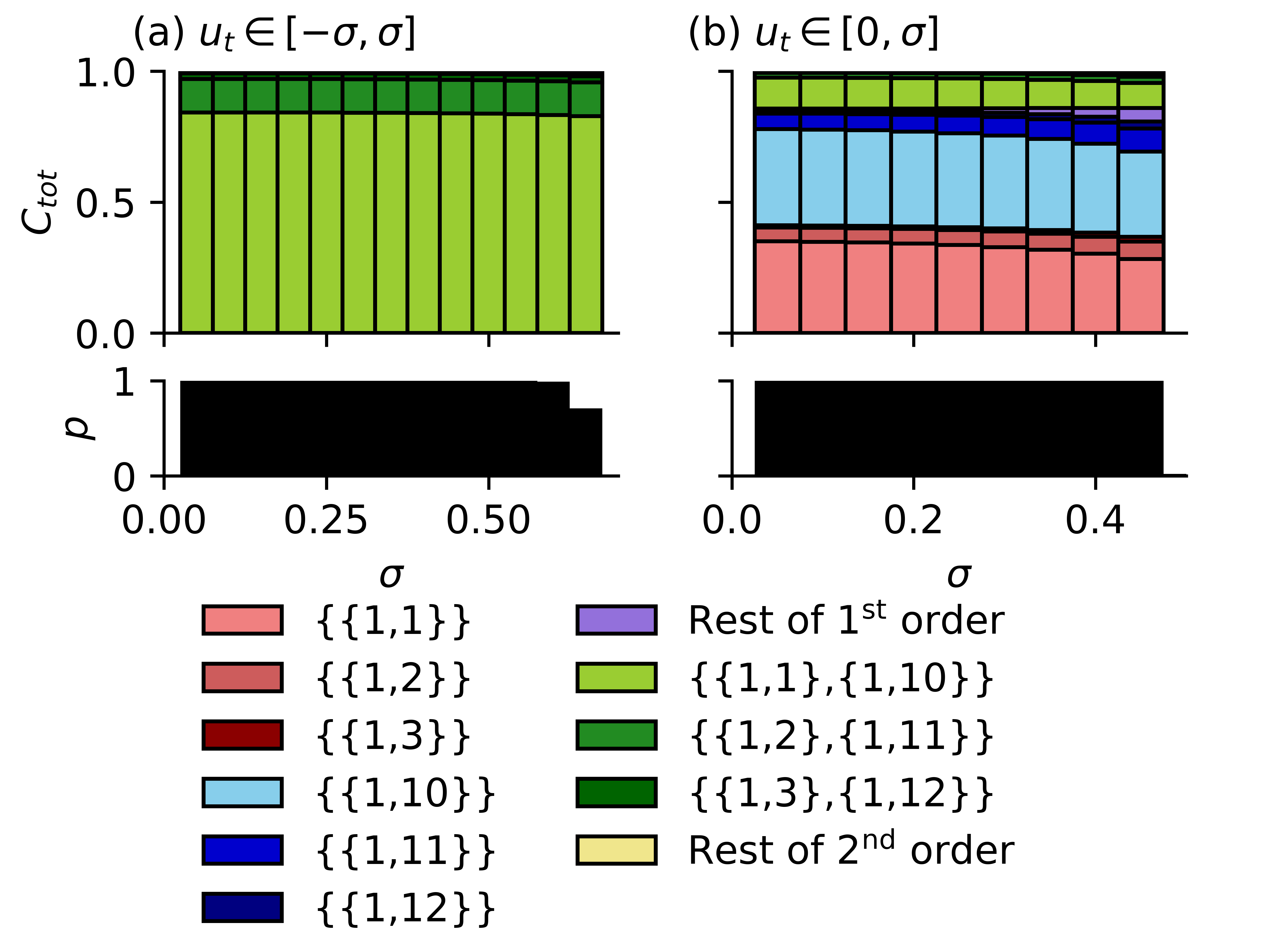}
    \caption{
        The IPC breakdown relative to $\sigma$ and the probability that $y_t$ does not diverge, $p$. 
        The labels indicate representative combinations of $\{\{n_s,s\}\}$, where $n_s$ is the degree of the polynomial, $s$ is the delayed time step of the input, and the desired output is $\prod_s P_{n_s}(\zeta_{t-s})$. 
        The labels for other combinations are omitted. 
        The probability $p$ represents the proportion of $y_t$ values that do not diverge to infinity and is calculated using 100 input time-series $\zeta_t$ generated from 100 random seeds. 
        In all the figures, the capacities are not stacked if the output diverged or $\sigma=0$. 
        In (a) and (b), the two ranges of $u_t$ are as follows: 
        (a) $u_t$ follows the uniform distribution in the range $[-\sigma,\sigma]$ and (b) in the range $[0,\sigma]$. 
    }
    \label{fig:ipc_narma10}
\end{figure}

\subsection{Application \#2: the machine learning network}
\begin{figure*}[tb]
    \centering
    \includegraphics[scale=0.85]{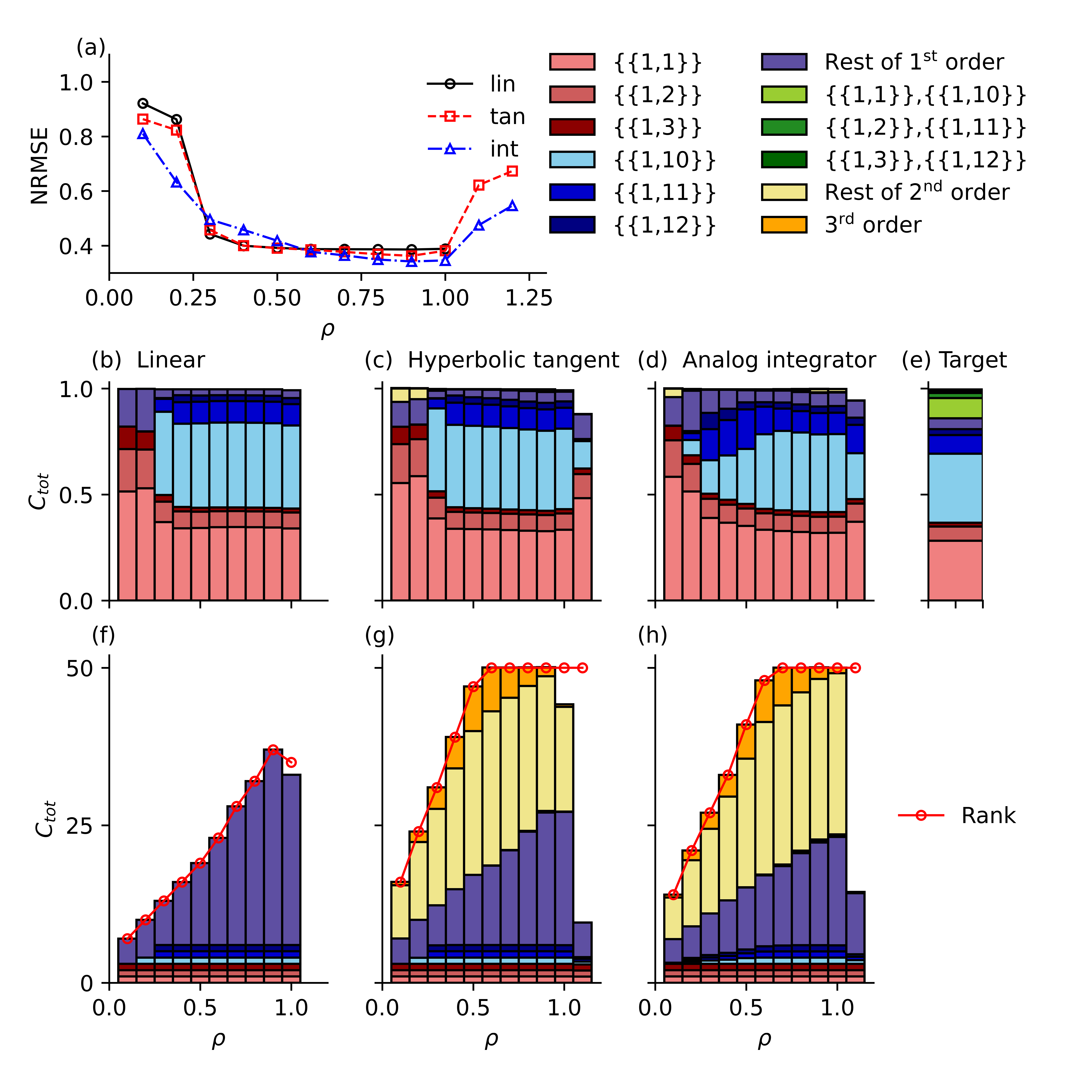}
    \caption{
        The performance breakdown of NARMA10-trained ESNs with the IPC. 
        Frame (a) shows the NRMSE of the NARMA10 task relative to the spectral radius $\rho$ of ESNs composed of linear, hyperbolic tangent, or analog integrator functions. 
        Panels (b), (c), and (d) display the relationship between $\rho$ and the IPC breakdown of the output of the NARMA10-trained ESN with linear, hyperbolic tangent, and analog integrator functions, respectively. 
        The labels indicate representative combinations of $\{\{n_s,s\}\}$, where $n_s$ is the degree of the polynomial, $s$ is the delayed time step of the input, and the desired output is $\prod_s P_{n_s}(\zeta_{t-s})$.
        In the case of first-order IPCs, the representative combinations---\{\{1,1\}\}, \{\{1,2\}\}, \{\{1,3\}\}, \{\{1,10\}\}, \{\{1,11\}\}, and \{\{1,12\}\}---are shown, and other combinations are summarized as ``rest of 1$^{\rm st}$ order''. 
        The representative combinations of second-order IPCs---\{\{1,1\},\{1,10\}\}, \{\{1,2\},\{1,11\}\}, and  \{\{1,3\},\{1,12\}\}---are barely held by the ESN, and thus, other combinations are summarized as ``rest of 2$^{\rm nd}$ order''. 
        Frame (e) shows the IPC breakdown of the target NARMA10 model with input in the $[0,0.45]$ range. 
        Panels (f), (g), and (h) represent the relationship between $\rho$ and the IPC breakdown of the ESN state with linear, hyperbolic tangent, and analog integrator functions, respectively. 
        The red line denotes the rank of state $r$, which corresponds to the total IPC in the time-invariant domain. 
    }
    \label{fig:performance_breakdown}
\end{figure*}

Second, we analyzed the performance of a machine learning network that solved the benchmark task using the ESN as an example.
As shown in FIG. \ref{fig:performance_breakdown}(a), we also emulated target NARMA10 model using 50-node ESNs with three activation functions---linear, hyperbolic tangent, and analog integral functions---and compared the output of the ESN $\hat{y}_t$ and the target $y_t$ with the normalized root-mean-square errors (NRMSEs). 
For all the functions, the NRMSE decreased as the spectral radius $\rho$ increased and increased when $\rho\ge 1$. 
To analyze the outputs of the ESNs training the NARMA10 model, the IPCs of the output $\hat{y}_t$ were estimated. 
FIG. \ref{fig:performance_breakdown}(b)--(d) shows the change in the IPC breakdown of the output $y_t$ with the increase in the spectral radius $\rho$ of the ESN with the linear, hyperbolic tangent, and analog integral functions, respectively.
To emulate the NARMA10 model, whose IPC breakdown is shown in FIG. \ref{fig:performance_breakdown}(e), the nine types of Legendre-chaoses $\{P_1(\zeta_{t-s})\}_{s=1,2,3,10,11,12}$ and $\{P_2(\zeta_{t-s})\}_{s=1,2,3}$ are mainly required in a certain ratio. 
According to FIG. \ref{fig:performance_breakdown}(b)--(d), for any activation function, as $\rho(<1)$ increases, $\{P_{1}(\zeta_{t-s})\}_{s=1,2,3,10,11,12}$ approaches the required rate. 
However, the three types of second-order IPCs $\{P_1(\zeta_{t-s})P_1(\zeta_{t-s-9})\}_{s=1,2,3}$ are almost zero in ESNs with an activation function. 
Therefore, in the NARMA10 task with ESNs, performance is compared using only the first-order IPCs.

To investigate why the second-order IPCs did not appear in the breakdown, we estimated the IPCs from ESN states. 
FIG. \ref{fig:performance_breakdown}(f)--(h) shows the change in the IPC breakdown with the increase in the spectral radius $\rho$ of the ESN with a linear, hyperbolic tangent and an analog integral function, respectively. 
The rank $r$ of the ESN state increases with $\rho(\le1)$ and corresponds to $C_{tot}$. 
As these three breakdowns have low-target second-order IPCs (each capacity is less than 0.2), the ESNs cannot emulate second-order Legendre-chaoses.

The above results demonstrate that our method is capable of decomposing the computational capability of the machine learning network before and after training, as well as clarifying whether the computational components required for the task have been extracted through learning and exist in the original network.

\subsection{Application \#3: the real neural network}
\begin{figure*}[tb]
    \centering
    \includegraphics[scale=0.7]{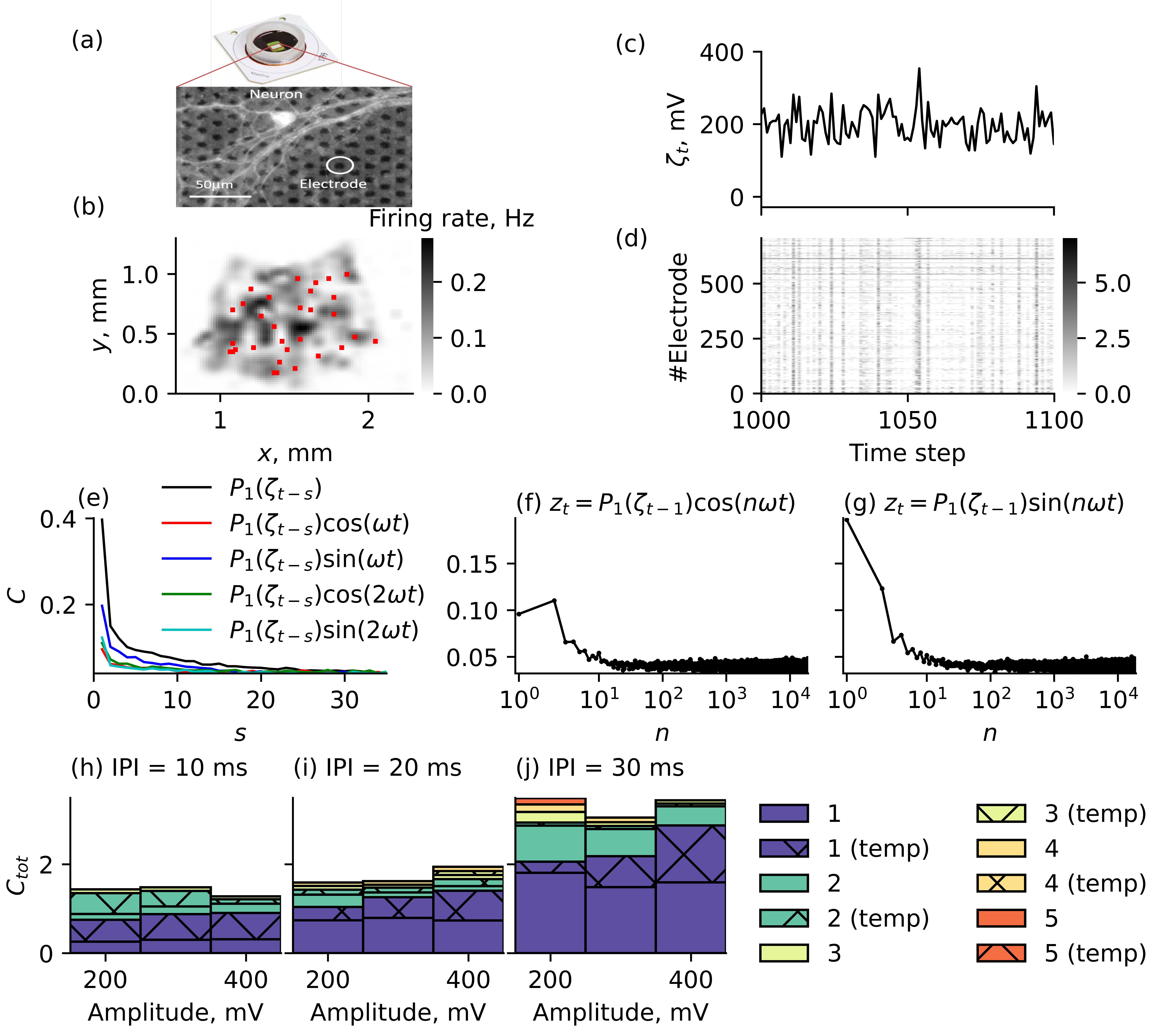}
    \caption{
        TIPCs of the dissociated culture of neurons. 
        Frame (a) shows the dissociated culture of neurons on the electrode array. 
        Panel (b) shows the spontaneous firing rate of the culture on the electrode array, where $(x,y)$ denotes the position of electrode. 
        The rate is filtered by the Gaussian kernel. 
        The red dots represent the 29 stimulation electrodes. 
        Frames (c) and (d) represent the amplitude of stimulus $\zeta_t$, which follows a Gaussian distribution ($\mu=200$ mV), and the spike count of each electrode ${\bm x}_t$ ($\mu=200$ mV and IPI = 10 ms), respectively. 
        Frame (e) illustrates the temporal memory functions (TMFs) of delay step $s$. 
        (e)--(g) show the TMFs when IPI = 30 ms and $\mu$ = 400 mV. 
        TMFs with a time-invariant target $z_t=P_1(\zeta_{t-1})$ and four time-variant targets---i.e.,  $z_t=P_1(\zeta_{t-1})\cos(n\omega t)$ and $P_1(\zeta_{t-1})\sin(n\omega t) ~ (n=1,2)$---are plotted. 
        Panels (f) and (g) show the first-order TIPC spectrum with the cosine $\zeta_t=P_1(\zeta_{t-1})\cos(n\omega t)$ and sine target $\zeta_t=P_1(\zeta_{t-1})\sin(n\omega t)$, respectively. 
        Frames (h), (i), and (j) depict the total capacity of the culture with IPI = 10, 20, and 30 ms, respectively, and the horizontal axis is the mean of amplitude $\mu$. 
        The hatched bar represents the total capacity with time-varying $n^{\rm th}$ order polynomials. 
    }
    \label{fig:culture_tipc}
\end{figure*}

Finally, to show the broad applicability of our theory, we prepared a dissociated culture of neurons for a physical reservoir, which is an open non-equilibrium system that fluctuates due to external inputs and has parameters that can be considered time-dependent. 
Real neurons extracted from the cortices of rat embryos were pharmacologically and mechanically isolated and then seeded on an electrode array. 
After the culture matured (FIG. \ref{fig:culture_tipc}[a]), we constructed a physical reservoir using electrodes with active neurons (FIG. \ref{fig:culture_tipc}[b]; see the Appendix). 
We repeatedly applied bipolar waves with a 10, 20, or 30 ms interpulse interval (IPI) to 29 stimulus electrodes, whose amplitude $\zeta_t$ follows a Gaussian distribution with mean $\mu=$ 200, 300, or 400 mV and standard deviation $\sigma=50$ mV (FIG. \ref{fig:culture_tipc}[c]). 
Furthermore, we computed the number of spikes in an IPI-width bin from $N=792$ measurement electrodes as the reservoir states (FIG. \ref{fig:culture_tipc}[d]). 
As a result, we obtained a long single trajectory of the activation of the electrodes according to the input stream, which consisted of 20,000 time steps and was used for our TIPC analysis.

Using these data, we computed the TIPCs of the physical system. 
FIG. \ref{fig:culture_tipc}(a) shows the first-order TIPCs of the delay step $s$, which contain the memory function (MF) \cite{jaeger2001short,white2004short} and the four temporal memory functions (TMFs), which had time-varying target $z_t=P_1(\zeta_{t-s})\cos(n\omega t)$ or $z_t=P_1(\zeta_{t-s})\sin(n\omega t)$ $(n=1,2)$. 
Note that $n\omega=2\pi n/T~(n=1,\ldots,T/2)$ denotes the frequency. 
The TMFs monotonically decay with an increase in $s$, as well as the MF. 
Next, to investigate the frequency characteristics of the TIPC, we plotted the TIPC with $s=1$. 
FIG. \ref{fig:culture_tipc}(b) and (c) shows the relationship between the frequency and TIPCs with the time-varying cosine and sine targets, respectively. 
We term such a graph the TIPC spectrum. 
Both spectra have larger TIPCs at lower frequencies. 
Therefore, the input was processed by the coupling terms of the low frequency sinusoidal wave and the past input, suggesting that the superposition of these waves represents a gradual trend---which may be caused, for example, by synaptic plasticity and neural adaptation---and thus, information processing was also embedded in the non-stationary changes.

FIG. \ref{fig:culture_tipc}(d)--(f) illustrates the total capacities $C_{tot}$ with different $\mu$ and IPI. 
Every $C_{tot}$ contained time-varying IPCs, and as the IPI decreased, the ratio of time-varying IPC to $C_{tot}$ increased. 
As with the time-invariant case, the time-variant IPC increased as the degree decreased. 
Therefore, the total capacity of the dissociated culture of neurons contained time-varying IPCs in all cases. 
This type of information processing could not be elucidated by the conventional IPC, suggesting that our proposed measure is effective.

\section{\label{sec:discussion}Discussion}

\subsection{The relationship between an attractor and information processing}
In the present paper, random variables were given as input to the time-invariant systems, whose states were represented by echo functions, which depend not on the time $t$ but only on the past input time-series. 
In the case where the system with no input converges to a fixed-point (e.g., the intersection point of the NARMA10 system), this function illustrates that the system stays at a fixed-point attractor and fluctuates around the fixed-point due to the input. 
The conventional IPC targets time-invariant systems and quantifies the input processing of a state that depends only on the input time-series. 
Therefore, the conventional IPC sometimes represents the input information processing performed around a certain fixed-point attractor.

In addition, the IPC is the squared norm of the coefficient vector of the temporal basis vectors expanded with PCs. 
Since these coefficient vectors obviously change depending on the fixed-point, different types of information processing are performed at different fixed points. 
However, the IPC extended for time-variant systems was applied to the limit cycle system, which does not satisfy the prerequisites for RC. 
From its TIPC estimates, the expanded solution contains the coupling terms of the input and the time-varying terms,  $P_1(\zeta_{t-s_1})\cos(\omega\tau t+\alpha_{1,s_1})$,  $P_2(\zeta_{t-s_1})\cos(\omega\tau t+\alpha_{2,s_1})$, and  $P_1(\zeta_{t-s_1})P_1(\zeta_{t-s_2})\cos(\omega\tau t+\alpha_{2,s_1,s_2})$, showing that the processed input represents the amplitude scale of the sinusoidal. 
Therefore, the fluctuation of a periodic attractor determined by input can be processed around the limit cycle.

Based on these findings, we conclude that the conventional IPC can evaluate computational capabilities around a fixed point, while the TIPC can also evaluate capabilities around a periodic attractor. 
Since the recent RC framework exploits not only fixed-points or periodic attractors but also chaotic ones \cite{sussillo2009generating,laje2013robust,nicola2017supervised,inoue2020designing}, future work should examine the relationship between various attractors and information processing.

\subsection{Methods to utilize a time-variant system as a computational resource}
In demonstrating the TIPCs for a limit cycle system, we showed that the state of the system can include coupling terms of the past input time-series. 
Since the coupling term is represented by the product of the time-varying basis $\psi_t$ and PC $\phi_t$, the conventional IPC for time-invariant terms could not quantify the amount of information. 
The TIPC indicates that even in a time-variant system, information can be processed by the PC $\phi_t$ in the coupling term. 
In addition, to date, RC has trained a static readout weight by linear regression, assuming that the ESP or FMP is satisfied; the state in the reservoir is a function of the input time-series. 
Since the target output is described by the input history, performance decreases when a time-dependent reservoir (e.g., an ESN with a spectral radius of $\rho\ge1$) is used. 
As the number of input history terms in the expanded state decreases, the performance drops, while the number of coupling terms increases. 
The input information in a coupling term can be used for the task and can be extracted by giving a readout weight that cancels out $\phi_t$ in the coupling term (e.g., a time-varying weight). 
Therefore, even in a reservoir where the ESP or FMP does not hold, the input time-series may be processed by TDPCs, and the task can be successfully solved by using new types of readout.

\begin{figure}[tb]
    \centering
    \includegraphics[scale=0.85]{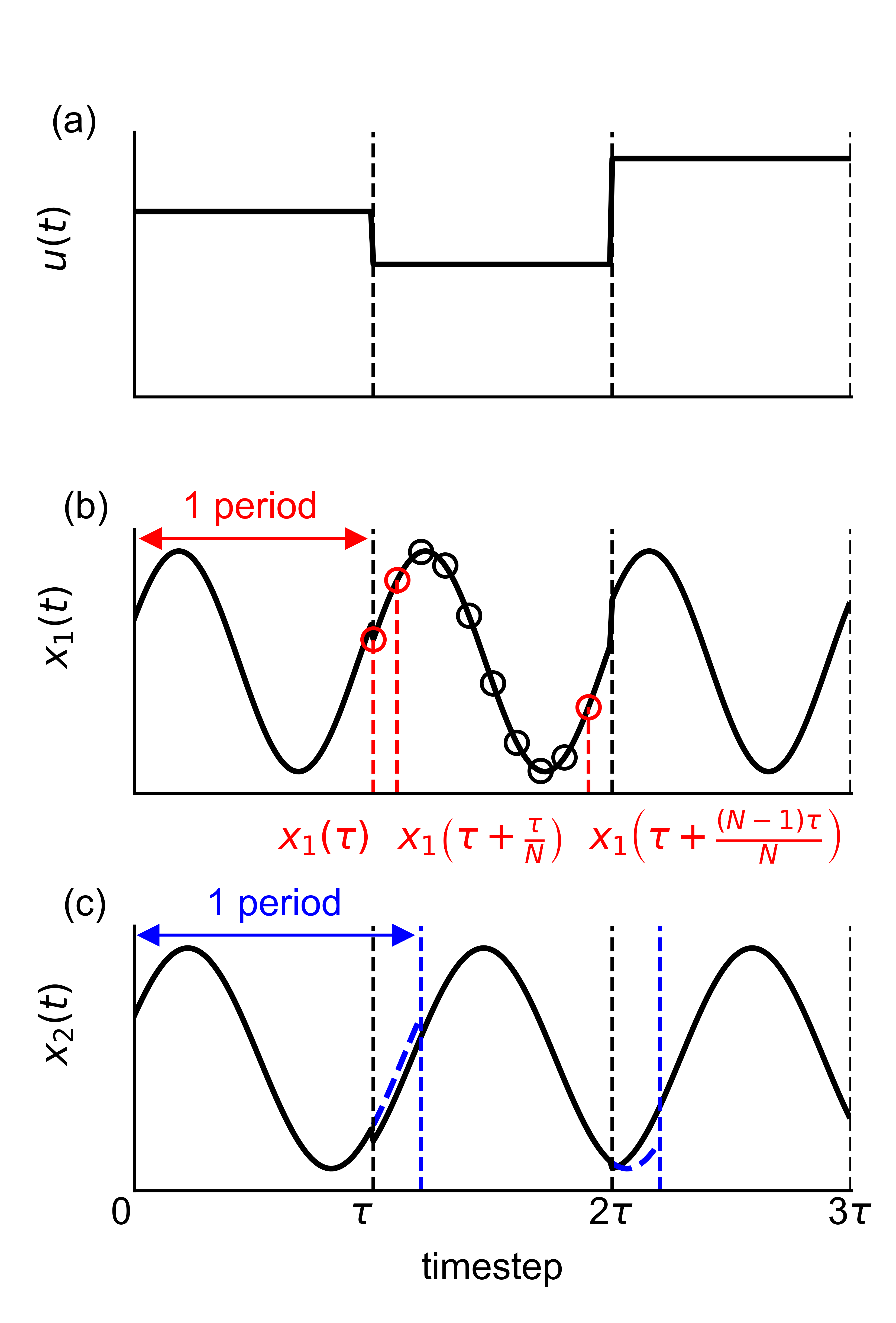}
    \caption{
        The time-multipleixing method transforms a periodic system into a time-invariant one. 
        (a) The input time-series $\{u_t\}$ is switched with time width $\tau$. 
        Frames (b) and (c) show a periodic state time-series whose frequency is and is not consistent with that of the input, respectively. 
        In (b), the circles represent virtual nodes ${\bm x}_1(t) = \left[x_1\left(t\right),x_1\left(t+\frac{\tau}{N}\right),\ldots,x_1\left(t+\frac{(N-1)\tau}{N}\right)\right]^\top$. 
    }
    \label{fig:time_multiplexing}
\end{figure}

In this connection, a method already exists for exploiting periodic systems as computational resources. 
As shown in FIG. \ref{fig:time_multiplexing}, the time-multiplexing technique \cite{appeltant2011information} switches the input $u_t$ with the time width $\tau$ (FIG. \ref{fig:time_multiplexing}[a]) and extracts ${\bm x}(t)=[x(t+\tau),x(t+2\tau),\ldots,x(t+(N-1)\tau)]^\top$ to virtually increase the number of nodes in the reservoir and improve the computational capabilities. 
If $x(t)$ is a periodic function and oscillates with a period specific to $u(t)$ (FIG. \ref{fig:time_multiplexing}[b]), applying an input with the same width $\tau$ as the period can extract time-invariant virtual nodes because the $i^{\rm th}$ virtual node always corresponds to $x(t)$ at a certain phase and is not affected by the periodic fluctuation. 
However, if the period does not match $\tau$ (FIG. \ref{fig:time_multiplexing}[c]), the phase shifts, and the scheme cannot exploit the computational capabilities of the periodic system. 
Therefore, time-multiplexing can be interpreted as a method capable of extracting a computational resource by transforming a time-varying system into a time-invariant system. 
Thus, a time-variant system can process input information through coupling terms, 
and rich input information can be extracted from time-variant systems by designing readout for the systems. 
In the future, information processing using a time-varying reservoir and novel design methods for readouts will be investigated.

\subsection{Extension to TIPC with Multiple Input Variables}
In the present paper, TIPC was limited to one type of input, but it could be easily extended to a multiple input version. 
Let $M$ independent stochastic inputs that follow multiple arbitrary distributions be $\zeta_t^{(1)},\ldots,\zeta_t^{(M)}\in\mathbb{R}$, and the state equation in Eq. (\ref{eq:x}) is modified as ${\bm x}_{t+1}={\bm f}({\bm x}_t,\zeta_{t}^{(1)},\ldots,\zeta_{t}^{(M)})$, whose state for a time-invariant system can be expanded by aPCs with multiple input variables \cite{ahlfeld2016samba}. 
As all of these aPCs are orthogonal, and their orthogonality is defined by the same inner product as Eqs. (\ref{eq:inner_product_re1}) and (\ref{eq:inner_product_re2}), which are also common to TDPCs, it is clear that we can expand the state for a time-variant system using the TDPCs and define its TIPCs. 
Physical systems often receive various inputs from the external environment, resulting in such multiple input-driven systems.
For example, in the dissociated culture of neurons, the ranks of the state are 724--792, but the total capacities are less than 3.5 (FIG. \ref{fig:culture_tipc}[c]), which are much smaller than their ranks. 
One possible speculation regarding this issue is that in our scheme, the state may be expressed as a function of multiple stochastic inputs, including the one we gave (e.g., electrical noise, synaptic noise, thermal noise, and shot noise \cite{faisal2008noise}). 
Of all the inputs, the ones we could observe is limited, 
and the total capacity computed only from the inputs did not reach the rank. 
Therefore, physical systems can receive unobservable inputs, which disturb the examination of all capacities.

\section{\label{sec:conclusion}Conclusion}
This paper attempted to clarify the unknown relationship between the PC expansion and IPC by deriving the IPC from the PC-expanded system. 
To illustrate this relation, we showed that the IPC can be measured using gPCs and aPCs. 
In addition, using the NARMA10 model, we concretely described the relationship and showed the usefulness of our theory. 
Next, taking into account the characteristics of the general solution of the input-driven dynamical system, we proposed the IPC for time-variant systems---called the TIPC. 
To demonstrate that the time-variant system has such TIPCs, we investigated the TIPC breakdown of the limit cycle system. 
The primary results are summarized as follows: 
\begin{itemize}
    \item Using SVD, we can obtain the orthogonal temporal basis vectors from the state time-series. 
    These vectors can be expanded with PCs to obtain the coefficient vector of each basis. 
    The IPC is equivalent to the squared norm of the coefficient of the PC used as the target output. 
    Therefore, the expanded basis coefficients represent the amount of input processing information.
    \item Using eight types of polynomials in the Askey scheme and Gram-Schmidt PCs, we estimated the IPCs of a one-dimensional ESN, whose total IPCs were equal to one. 
    These results indicate that various types of PCs within the Askey scheme and Gram-Schmidt orthogonalization are suitable for the target output of the IPC. 
    \item We calculated the TIPC of a time-variant system using the simple limit cycle and revealed that the input information processing is performed by time- and input time-series-dependent terms. 
    \item IPC analysis revealed that the NARMA10 model is mainly composed of $P_1(\zeta_{t-s})~(s=1,2,3,10,11,12)$ and $P_1(\zeta_{t-s})P_1(\zeta_{t-s-9})~(s=1,2,3)$. 
    The NARMA10 benchmark task can be solved by holding the nine types of input information in a reservoir. 
    Consequently, combining the IPC and PC expansion provides a clear and concise picture of information processing. 
    \item 
    The dissociated culture of neurons had not only the time-invariant IPC but also the time-variant IPC, suggesting that the TIPC reveals the information processing in a trend---for example, synaptic plasticity and neural adaptation---that has been left behind to date. 
\end{itemize}
The above results suggest that the connection between the IPC and PC expansion allows for a simpler description of information processing in dynamical systems. 
In the future, information processing using time-variant systems should also be elucidated. 
This scheme can be applied to non-stationary systems and thus may be suitable for elucidating information processing in neural circuits which has been overlooked so far. 
In addition, it can be applied not only to the neural systems but also to other physical systems that can be time-variant---for example, fluid, quantum, spintronics, and optical systems. 
For example, recently it is reported that some types of vortex generated when a fluid flows past a bluff body can be used as an information processing device \cite{goto2020computing}. 
In their analysis, they found that near the critical Reynolds number, where the flow exhibits a twin vortex before the onset of the Karman vortex shedding associated with the Hopf bifurcation, the information processing capability was maximized. 
This was also characterized by the breakdown of ESP. 
According to our results, it may be possible to evaluate the type and amount of information processing even in the Karman vortex shedding, which would be a direction for future work.

\begin{acknowledgments}
This paper is based on results obtained from the Exploration of Neuromorphic Dynamics towards Future Symbiotic Society project commissioned by NEDO, KAKENHI grant (17K20090), AMED (JP18dm0307009) and Asahi Glass Foundation.
H. T. was partly supported by JSPS KAKENHI (20H04252), AMED (JP20dm0307009),  the Naito Foundation, and the Asahi Glass Foundation. 
K. N. was supported by the JSPS KAKENHI Grant Number JP18H05472 and MEXT Quantum Leap Flagship Program (MEXT Q-LEAP) Grant Number JPMXS0118067394.
\end{acknowledgments}

\appendix

\renewcommand{\theequation}{A\arabic{equation}}
\setcounter{equation}{0}
\renewcommand{\thetable}{A\arabic{table}}
\setcounter{table}{0}
\renewcommand{\thefigure}{A\arabic{figure}}
\setcounter{figure}{0}

\section{\label{sec:Polynomial}An ESN and LSM}
In an ESN composed of $N$-nodes, the $i^{\rm th}~(i=1,2,\ldots,N)$ node state $x_{i,t}$ at the $t^{\rm th}$ time step can be written as follows:
\begin{eqnarray}
    x_{i,t+1} &=& f\left(\sum_{j=1}^N w_{ij}x_{j,t} + w_{in,i}u_t\right), 
\end{eqnarray}
where $f$ is the activation function, and $w_{ij}$ and $w_{in,i}$ are the internal and input weights, respectively. 

To emulate the target output $z_t$, we use linear regression to obtain an estimate of $z_t$, $\tilde{z}_t$, as follows:
\begin{eqnarray}
    \tilde{z}_t &=& \tilde{\bm w}_{out}^\top\cdot{\bm x}_t, \label{eqA:ztilde}\\
    \tilde{\bm w}_{out} &=& \arg \min_{{\bm w}_{out}}\sum_{t=1}^T \left(z_t-{\bm w}_{out}^\top\cdot{\bm x}_t\right)^2, \label{eqA:wtilde}
\end{eqnarray}
where ${\bm w}_{out}$ and $\tilde{\bm w}_{out}\in\mathbb{R}^L$ are the weight and solution vector for the target output, respectively. 
This learning method does not affect the state of the reservoir but places a constraint on ${\bm x}_t$. 
Jaeger \cite{jaeger2004harnessing} and Maass \cite{maass2002real} independently developed RC by integrating ESNs and LSMs, respectively. 
The prerequisites for ${\bm x}_t$ differ in ESNs and LSMs.

An ESN requires ${\bm x}_t$ to be an echo function, which is a function of only the past input time-series ${\bm u}_t=\{u_{t-s}\}_{s=1}^t$. 
This dynamical property is referred to as the ESP \cite{jaeger2002tutorial,yildiz2012re,manjunath2013echo}. 
We examine this feature by giving the same input time-series to two systems with different initial values and checking whether the two states coincide after a sufficiently long period of time. 
If the input is noise, this phenomenon is called noise-induced synchronization \cite{maritan1994chaos,toral2001analytical}. 
Furthermore, when the input is generated from a deterministic system, the phenomenon wherein the state is synchronized with the input is called generalized synchronization \cite{lu2018attractor}. 
Therefore, ESP is related to the synchronization phenomenon of nonlinear dynamical systems.

In addition, LSMs impose a prerequisite on the power series expansion of states. 
If the system is time-invariant---i.e., the state does not depend on time---and retains exponentially decaying inputs, its state can be approximated by the Volterra series \cite{boyd1985fading}, which is a series expansion with non-orthogonal bases. 
Accordingly, the Volterra series operator \cite{volterra1959theory} can be expanded in a non-orthogonal power series, as follows: 
\begin{eqnarray}
    {\bm x}_t = \sum_{n=1}^\infty \prod_{s_1=1}^{t}\prod_{s_2=s_1}^{t}\cdots\prod_{s_n=s_{n-1}}^{t} {\bm g}_{s_1\cdots s_n}u_{t-s_1}\cdots u_{t-s_n}, \nonumber \\
\end{eqnarray}
where 
${\bm g}_{s_1\cdots s_n}\in\mathbb{R}^L$ is the $n^{\rm th}$ Volterra kernel. 
Such a memory decay feature is called the FMP \cite{maass2002real,maass2004computational,maass2011liquid} and is considered a prerequisite for LSMs.

\renewcommand{\theequation}{B\arabic{equation}}
\setcounter{equation}{0}
\renewcommand{\thetable}{B\arabic{table}}
\setcounter{table}{0}
\renewcommand{\thefigure}{B\arabic{figure}}
\setcounter{figure}{0}

\section{\label{sec:Polynomial}Polynomial chaos expansion}
A deterministic dynamical system with a stochastic input $\zeta_{t}$ can be described as an operator of $\{\zeta_{t-s}\}_{s=1}^\infty$ according to polynomial chaos expansion \cite{wiener1938homogeneous}, 
which is a series expansion using the target outputs of IPC as the bases---i.e., multivariate orthogonal polynomials of the random variables, $z_t^{(i)}$, described in Eq. (\ref{eq:z}). 
These multivariate polynomials are referred to as polynomial chaoses (PCs), and the space spanned by the PCs is called homogeneous chaos \cite{wiener1938homogeneous}, expressed as 
\begin{eqnarray}
    {\bm x}_t = \sum_{i=1}^{\infty} {\bm c}_i z_t^{(i)}, \label{eq:expanded_x}
\end{eqnarray}
where ${\bm c}_i\in\mathbb{R}^L$ is the $i^{\rm th}$ coefficient vector.

\begin{figure*}[tb]
    \centering
    \includegraphics[scale=0.85]{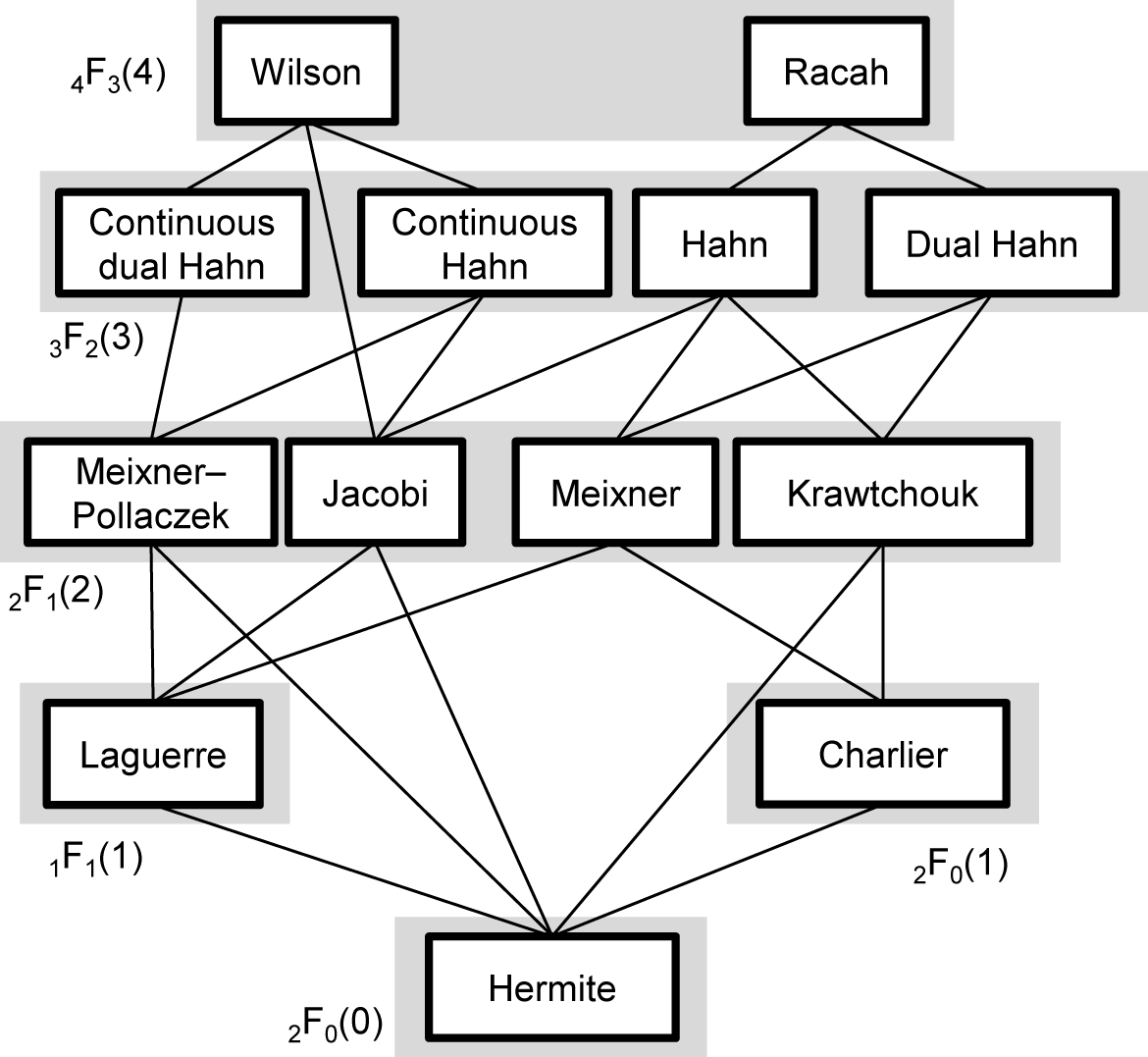}
    \caption{
        The relationship between polynomials in the Askey scheme. 
        Each polynomial is represented with the hypergeometric series $_rF_s(p)$, where $p$ represents the number of parameters substituted in $a_i~(i=1,2,\ldots,r)$, $b_i~(i=1,2,\ldots,s)$, or $z$. 
        An upper polynomial with the limit of a certain parameter or parameters becomes a lower polynomial connected with a line. 
    }
    \label{figA:askey_scheme}
\end{figure*}

\begin{table*}[tb]
    \begin{center}
        \caption{
            Sets of random variables and polynomial chaoses 
        }
        \begin{tabular}{c|c|c|c} \hline
            \multicolumn{2}{c|}{Support of random variable} & Random variable $\zeta_t$ & Polynomial chaos $z_t^{(i)}$ \\ \hline \hline
  Continuous & $(-\infty,\infty)$ & Gaussian & Hermite-chaos  \\
            & $[0,\infty)$ & Gamma & Laguerre-chaos \\
            & $[a,b]$ & Beta & Jacobi-chaos \\
            & $[a,b]$ & Uniform & Legendre-chaos \\ \hline
    Discrete & $\{0,1,\ldots\}$ & Poisson & Charlier-chaos \\
            & $\{0,1,\ldots,N\}$ & Binomial & Krawtchouk-chaos \\
            & $\{0,1,\ldots\}$ & Negative binomial & Meixner-chaos \\
            & $\{0,1,\ldots,N\}$ & Hypergeometric & Hahn-chaos \\ \hline
        \end{tabular}
    \label{tabA:gPC}
  \end{center}
\end{table*}

\subsection{Generalized polynomial chaos}
Generalized polynomial chaos (gPC) \cite{xiu2002wiener} is PC composed of the univariate polynomial $\mathcal{F}_n(\zeta)$ included in the Askey scheme \cite{askey1985some} tree (FIG. \ref{figA:askey_scheme}). 
The Askey scheme represents various orthogonal polynomials (e.g., Hermite, Jacobi, Laguerre, and Charlier) using the hypergeometric series $_rF_s$ of $x$, along with parameters $a_1,\ldots,a_r$ and $b_1,\ldots,b_s$:
\begin{eqnarray}
    _rF_s\left(x;
    \begin{matrix}
        a_1,\ldots,a_r \\
        b_1,\ldots,b_s \\
    \end{matrix}
    \right) = \sum_{k=0}^\infty \frac{(a_1)_k\cdots(a_r)_k}{(b_1)_k\cdots(b_s)_k} \frac{x^k}{k!}, \nonumber \\ \\
    (a)_n =
    \begin{cases}
        1 & (n=0) \\
        a(a+1)\cdots(a+n-1) & (n=1,2,\ldots)
    \end{cases},\nonumber\\
\end{eqnarray}
where $(a)_n$ is the Pochhammer symbol. 
For example, the Laguerre polynomial with a parameter $\alpha$, $L^{(\alpha)}_n(\zeta)$ can be written as
\begin{eqnarray}
    L^{(\alpha)}_n(\zeta) = \frac{(\alpha+1)_n}{n!} {_1F_1}\left( \zeta; 
    \begin{matrix}
        -n \\
        \alpha+1
    \end{matrix}
    \right). \nonumber
\end{eqnarray}
Note that in FIG. \ref{figA:askey_scheme}, an upper polynomial with the limit of a certain parameter or parameters becomes a lower polynomial connected with a line; for example, the Laguerre polynomial becomes the Hermite polynomial $H_n(\zeta)$ by taking the following limit of $\alpha$:
\begin{eqnarray}
    \lim_{\alpha\rightarrow\infty} \left( \frac{2}{\alpha} \right)^{n/2} L_n^{(\alpha)} \left( (2\alpha)^{1/2}\zeta + \alpha \right) = \frac{(-1)^n}{n!} H_n(\zeta). \nonumber
\end{eqnarray}

The $i^{\rm th}$ target $z^{(i)}_t$ is represented by the product of  the $n^{(i)}_s$-th order polynomial of $\zeta_t$ delayed by $k^{(i)}_s$ steps, $\mathcal{F}_{n^{(i)}_s}(\zeta_{t-k^{(i)}_s})$. 
When the sets of degree $n$ and delay step $k$ are given by the $i^{\rm th}$ family of sets $\mathcal{N}_i=\{(n_1^{(i)},k_1^{(i)}),(n_2^{(i)},k_2^{(i)}),\ldots\}$, the target output is represented as
\begin{eqnarray}
    z_t^{(i)} &=& \prod_{s} \mathcal{F}_{n_s^{(i)}} \left(\zeta_{t-k_s^{(i)}}\right). 
\end{eqnarray}
Using the hypergeometrical series $_rF_s(\zeta)$, $\mathcal{F}_{n}(\zeta_{t-k})$ can be applied to the following eight types of orthogonal polynomials.

\subsubsection*{The Hermite polynomial and a Gaussian distribution}
The $n^{\rm th}~(n=1,2,\ldots)$ order Hermite polynomial $H_n(\zeta)$ is given by 
\begin{eqnarray}
    H_{n+1} = \zeta H_n - nH_{n-1}, \label{eqA:Hermite_polynomial}
\end{eqnarray}
where $H_0=1$, and $H_{-1}=0$. 
$\zeta$ follows a standard normal distribution
\begin{eqnarray}
    f(\zeta)=\frac{1}{\sqrt{2\pi}}\exp\left(\frac{\zeta^2}{2}\right). \label{eqA:starndard_normal_dist}
\end{eqnarray}

\subsubsection*{The Laguerre polynomial and a gamma distribution}
The $n^{\rm th}~(n=1,2,\ldots)$ order Laguerre polynomial $L^{(\alpha)}_n(\zeta)$ is given by 
\begin{eqnarray}
    L^{(\alpha)}_n(\zeta) &=& \sum_{i=0}^n (-1)^i \left( 
    \begin{matrix}
        n+\alpha \\
        n-i
    \end{matrix}
    \right) \frac{\zeta^i}{i!}, \label{eqA:Laguerre_polynomial}
\end{eqnarray}
where $\zeta$ follows a gamma distribution
\begin{eqnarray}
    f(\zeta)=\frac{1}{\Gamma(\alpha+1)\beta^{\alpha+1}}\zeta^\alpha e^{-\zeta/\beta}. \label{eqA:gamma_dist}
\end{eqnarray}
The parameter $\alpha>-1$, $\beta=1$, and $\Gamma(\cdot)$ is the gamma function. 
In FIG. \ref{fig:theory_demo}(b), the parameter $\alpha$ was set to $1$.

\subsubsection*{The Jacobi polynomial and a beta distribution}
The $n^{\rm th}~(n=1,2,\ldots)$ order Jacobi polynomial $P^{(\alpha,\beta)}_n(\zeta):=P_n$ is given by 
\begin{eqnarray}
    P_{n+1} &=& 
    \frac{(\gamma_n+1)( \gamma_n(\gamma_n+2)\zeta+\alpha^2-\beta^2 )P_n}{2(n+1)(\gamma_n-n+1)\gamma_n} \nonumber\\
    &-& \frac{2(n+\alpha)(n+\beta)(\gamma_n+2)}{2(n+1)(\gamma_n-n+1)\gamma_n}P_{n-1}, 
\end{eqnarray}
where $\gamma = 2n+\alpha+\beta$, and 
$\zeta$ follows a beta distribution in the range of $[-1,1]$:
\begin{eqnarray}
    f(\zeta) &=& \frac{1}{B(\alpha,\beta)}\zeta^{\alpha-1} (1-\zeta)^{\beta-1}, \label{eqA:beta_dist} \\
    B(\alpha,\beta) &=& \int_{0}^1 t^{\alpha-1} (1-t)^{\beta-1}dt. 
\end{eqnarray}
In FIG. \ref{fig:theory_demo}(c), the parameters $(\alpha,\beta)$ were set to $(-0.25,-0.25)$.

\subsubsection*{The Legendre polynomial and a uniform distribution}
The $n^{\rm th}~(n=1,2,\ldots)$ order Legendre polynomial $P_n(\zeta)$ is given by 
\begin{eqnarray}
    P_n(\zeta) &=& \sum_{k=0}^{\lfloor n/2\rfloor} \frac{(-1)^{-k}}{2^n}
    \left(
    \begin{matrix}
        n \\
        k
    \end{matrix}
    \right) \left(
    \begin{matrix}
        2n-2k \\
        n
    \end{matrix}
    \right)
    \zeta^{n-2k}, \nonumber\\
    \label{eqA:Legendre_polynomial}
\end{eqnarray}
where $\lfloor\cdot\rfloor$ represents the floor function, and $\zeta$ follows a uniform distribution in the range of $[-1,1]$:
\begin{eqnarray}
    f(\zeta) &=& \frac{1}{2}. \label{eqA:uniform_dist} 
\end{eqnarray}

\subsubsection*{The Charlier polynomial and a Poisson distribution}
The $n^{\rm th}~(n=1,2,\ldots)$ order Charlier polynomial $C_n(\zeta; a)$ is given by 
\begin{eqnarray}
    C_n(\zeta;a) = \sum_{i=0}^n \left( 
    \begin{matrix}
        n \\
        i
    \end{matrix}
    \right) \left( 
    \begin{matrix}
        \zeta \\
        i
    \end{matrix}
    \right) i! (-a)^{n-i}, \label{eqA:Charlier_polynomial} 
\end{eqnarray}
where $\zeta$ follows a Poisson distribution
\begin{eqnarray}
    f(\zeta) &=& \frac{e^{-a}a^\zeta}{\zeta!}. \label{eqA:poisson_dist}
\end{eqnarray}
In FIG. \ref{fig:theory_demo}(e), the parameter $\alpha$ was set to $6$.

\subsubsection*{The Krawtchouk polynomial and a binomial distribution} 
The $n^{\rm th}~(n=1,2,\ldots)$ order Krawtchouk polynomial $K_n(\zeta; p,N)$ is given by 
\begin{eqnarray}
    K_{n}(\zeta;p,N) = 
    \sum_{i=0}^n (-1)^{n-i} \left(
    \begin{matrix}
        N-\zeta \\
        n-i
    \end{matrix}
    \right) \left(
    \begin{matrix}
        \zeta \\
        i
    \end{matrix}
    \right) p^{n-i} (1-p)^{i}, \nonumber\\ \label{eqA:Krawtchouk_polynomial} 
\end{eqnarray}
where $\zeta$ follows a binomial distribution
\begin{eqnarray}
    f(\zeta) &=& \left(
    \begin{matrix}
        N \\
        \zeta
    \end{matrix}
    \right) p^\zeta (1-p)^{N-\zeta}. \label{eqA:binomial_dist}
\end{eqnarray}
In FIG. \ref{fig:theory_demo}(f), the parameters $(p,N)$ were set to $(0.5,10)$.

\subsubsection*{The Meixner polynomial and a negative binomial distribution}
The $n^{\rm th}~(n=1,2,\ldots)$ order Meixner polynomial $M_n(\zeta;\beta,c):=M_n$ is given by 
\begin{eqnarray}
    M_{n+1} = \frac{\{(c-1)\zeta+n+c(n+\beta)\}M_n-nM_{n-1}}{c(n+\beta)}, \nonumber\\ \label{eqA:Meixner_polynomial}
\end{eqnarray}
where $M_0=1$ and $M_{-1}=0$. 
$\zeta$ follows a negative binomial distribution
\begin{eqnarray}
    f(\zeta;\beta,c) &=& \left(
    \begin{matrix}
        \zeta-1 \\
        \beta-1
    \end{matrix}
    \right) (1-c)^\beta c^{\zeta-\beta}. \label{eqA:negative_binomial_dist}
\end{eqnarray}
In FIG. \ref{fig:theory_demo}(g), the parameters $(c,\beta)$ were set to $(0.2,10)$.

\subsubsection*{The Hahn polynomial and a hypergeometric distribution}
The $n^{\rm th}$ order Hahn polynomial $Q_n(\zeta;\alpha,\beta,N):=Q_n$ is given by 
\begin{eqnarray}
    Q_{n+1} &=& \frac{A_n+C_n-\zeta}{A_n} Q_{n} - \frac{C_n}{A_n} Q_{n-1}, \label{eqA:Hahn_polynomial} \\
    A_n &=& \frac{(n+\alpha+\beta+1)(n+\alpha+1)(N-n)}{(2n+\alpha+\beta+1)(2n+\alpha+\beta+2)}, \\
    C_n &=& \frac{n(n+\alpha+\beta+N+1)(n+\beta)}{(2n+\alpha+\beta)(2n+\alpha+\beta+1)},
\end{eqnarray}
where $Q_0=1$ and $Q_{-1}=0$. 
$\zeta$ follows a hypergeometric distribution with $m=-\alpha-1$ and $n=-\beta-1$: 
\begin{eqnarray}
    f(\zeta;m,n,N) &=& \left(
    \begin{matrix}
        m \\
        \zeta
    \end{matrix}
    \right) \left(
    \begin{matrix}
        n \\
        N-\zeta
    \end{matrix}
    \right)p^\zeta (1-p)^{N-\zeta}.\nonumber\\ \label{eqA:hypergeometric_dist}
\end{eqnarray}
In FIG. \ref{fig:theory_demo}(h), the parameters $(m,n,N)$ were set to $(100,50,20)$.

\subsection{Arbitrary polynomial chaos}
In addition, arbitrary polynomial chaos (aPC) \cite{oladyshkin2012data} is the PC for random variables following an arbitrary probability distribution, 
and we can compute univariate polynomials by applying the Gram-Schmidt orthogonalization procedure. 
The $n^{\rm th}$ Gram-Schmidt polynomial $\mathcal{F}_n(\zeta_t)=\psi_t^{(n)}$ is obtained from the following equations: 
\begin{eqnarray}
    \psi_t^{(n)} &=& \zeta_t^n - \sum_{i=0}^{n-1} c_i^{(n)} \psi_t^{(i)}, \\
    c_i^{(n)} &=& \frac{({\bm\psi}^{(i)})^\top\cdot{\bm\zeta}^n}{({\bm\psi}^{(i)})^\top\cdot{\bm\psi}^{(i)}}, 
\end{eqnarray}
where ${\bm\zeta}^n=[\zeta_1^n\cdots\zeta_T^n]^\top$ and ${\bm\psi}^{(i)}=[\psi_1^{(i)}\cdots\psi_T^{(i)}]^\top$.
Note that $\psi_t^{(0)}=1$. 
From the univariate polynomial $\mathcal{F}_n(\zeta)$ and Eq. (\ref{eq:z}), we can calculate the polynomial chaos for an arbitrary input distribution, Gram-Schmidt polynomial chaos.

If the input $\zeta_t$ follows a certain distribution, PC, including gPC and aPC, is determined based on its orthogonality. 
If a weighting function $w({\bm\zeta})$ specific to PCs $\{z_t^{(i)}\}_{i=1}^\infty$ exists, 
the following orthogonality relations should be satisfied: 
\begin{eqnarray}
    \left< z_t^{(i)} z_t^{(j)} \right> &=& \left<(z_t^{(i)})^2\right>\delta_{ij}, \label{eqA:inner_product_z} \\
    \left< f({\bm\zeta})g({\bm\zeta}) \right> &=& \sum_{\bm\zeta} w({\bm\zeta})f({\bm\zeta})g({\bm\zeta}), \label{eqA:inner_product_define}
\end{eqnarray}
where $\delta_{ij}$ is the Kronecker delta function. 
Several sets of random variables and gPCs were proposed \cite{xiu2002wiener}, as shown in TABLE \ref{tabA:gPC}. 
Note that expansion, in terms of the Hermite-chaos of Gaussian variables and Charlier-chaos of Poisson variables, converges in the sense of $L_2$ according to the Cameron-Martin theorem \cite{cameron1947orthogonal} and Ogura \cite{ogura1972orthogonal}, respectively.

\renewcommand{\theequation}{C\arabic{equation}}
\setcounter{equation}{0}
\renewcommand{\thetable}{C\arabic{table}}
\setcounter{table}{0}
\renewcommand{\thefigure}{C\arabic{figure}}
\setcounter{figure}{0}

\section{\label{sec:threshold}Threshold of IPC}
To remove the estimation error, $C({\bm X},{\bm z})$ was set to zero if it was smaller than the threshold $\epsilon$: 
\begin{eqnarray}
    C_\epsilon({\bm X},{\bm z}) = \theta_\epsilon \left( C({\bm X},{\bm z})\right) C({\bm X},{\bm z}), 
\end{eqnarray}
where $\theta_\epsilon (\cdot)$ is the Heaviside step function. 
The threshold is determined using random shuffle surrogates. 
We prepared $N(= 200)$ surrogates that are $z_t$ shuffled in the time direction 
and calculated IPCs using the surrogates to obtain $N$ capacities. 
Furthermore, we let the significance level be $\alpha(=1)\%$ and chose the original IPC, which exceeds 1.2--3 times the value in the top $\alpha/2\%$ of $N$ capacities. 
The above operation was performed for each $z_t$, and significant IPCs were obtained.

\begin{figure*}[tb]
    \centering
    \includegraphics[scale=0.85]{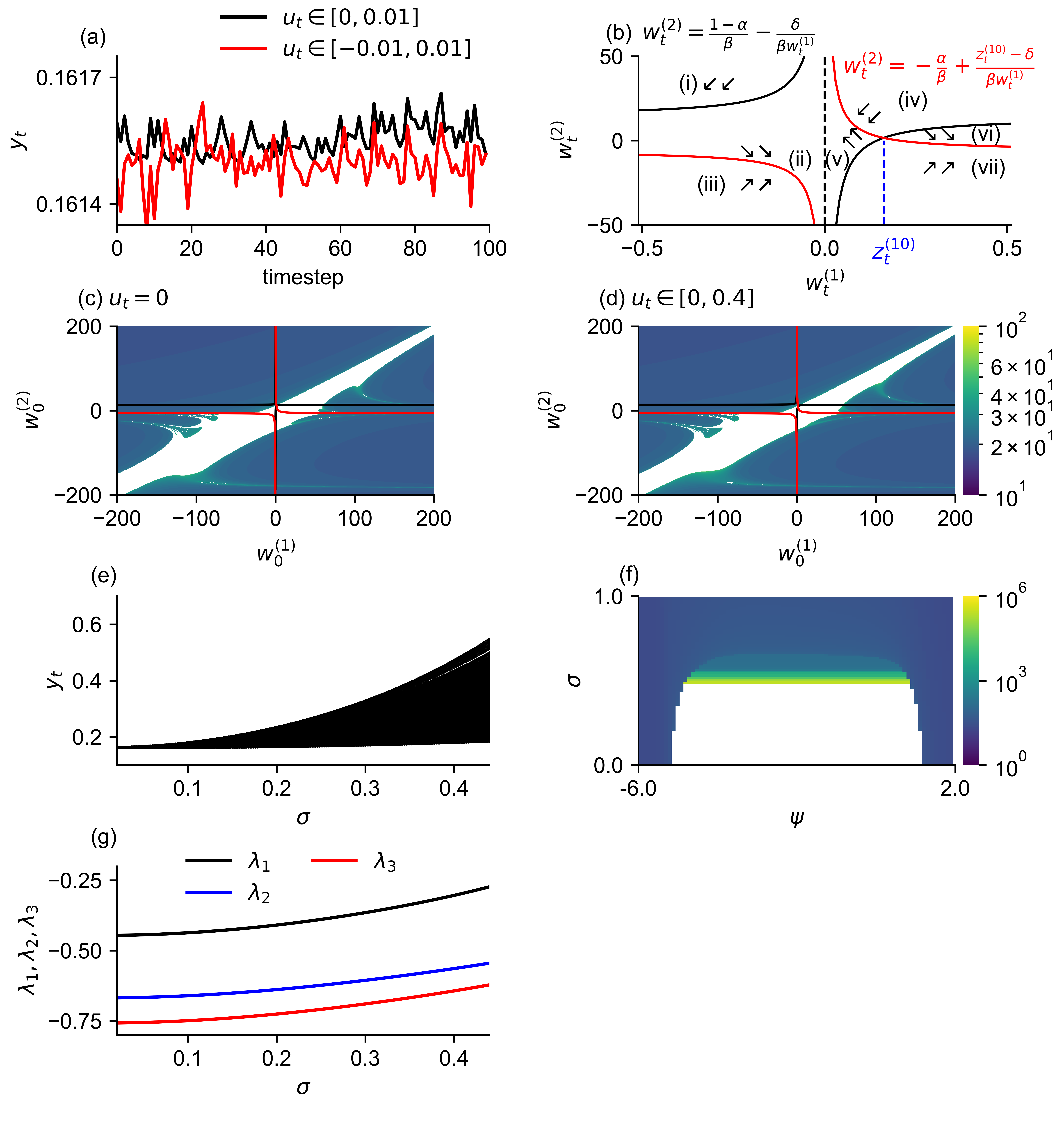}
    \caption{
        Properties of the NARMA10 dynamical system with $(\alpha,\beta,\gamma,\delta)=(0.3,0.05,1.5,0.1)$. 
        Panel (a) shows the time-series of Eq. (\ref{eq:narma10}) with $u_t\in[0,0.01]$ (black) and $u_t\in[-0.01,0.01]$ (red) and 
        (b) the stability diagram of $w^{(1)}_t=z_t^{(1)}$ and $w_t^{(2)}=\sum_{i=1}^{10}z_t^{(i)}$ as $z_t^{(10)}>\delta$. 
        The two nullclines, $w_t^{(2)}=\frac{1-\alpha}{\beta}-\frac{\delta}{\beta w_t^{(1)}}$ (solid black) and $w_t^{(2)}=-\frac{\alpha}{\beta}+\frac{z_t^{(10)}-\delta}{\beta w_t^{(1)}}$ (solid red), are displayed. 
        The signs of $\Delta w_t^{(1)}$ and $\Delta w_t^{(2)}$ are shown in TABLE \ref{tabA:1}. 
        Note that the point $(z_t^{(10)}, (1-\alpha)/\beta-\delta/\beta z_t^{(10)})$ is a saddle point. 
        Panels (c) and (d) show the basins of attraction of Eq. (\ref{eq:narma10}) relative to $w_0^{(1)}$ and $w_0^{(2)}$ with $u_t=0$ and $u_t\in[0,0.4]$, respectively, 
        as well as the two nullclines from (b). 
        In (c), (d), and (f), each dot shows the time step at which $y_t$ diverges to infinity, 
        and the white dot represents the initial values for which $y_t$ does not do so. 
        Frame (e) illustrates the bifurcation diagram of Eq. (\ref{eq:narma10}), 
        while (f) shows the basin of attraction of Eq. (\ref{eq:narma10}) relative to $\sigma$ and $\psi$, 
        where the initial values were $y_i=\psi~(i=0,1,\ldots,9)$. 
        The three largest Lyapunov spectra $\lambda_i~(i=1,2,3)$ relative to $\sigma$ are shown in (g). 
    }
    \label{figA:narma10_property}
\end{figure*}

\renewcommand{\theequation}{D\arabic{equation}}
\setcounter{equation}{0}
\renewcommand{\thetable}{D\arabic{table}}
\setcounter{table}{0}
\renewcommand{\thefigure}{D\arabic{figure}}
\setcounter{figure}{0}

\section{Classification of the NARMA10 model}
\subsection{Attractor analysis}
\begin{figure*}[tb]
    \centering
    \includegraphics[scale=0.85]{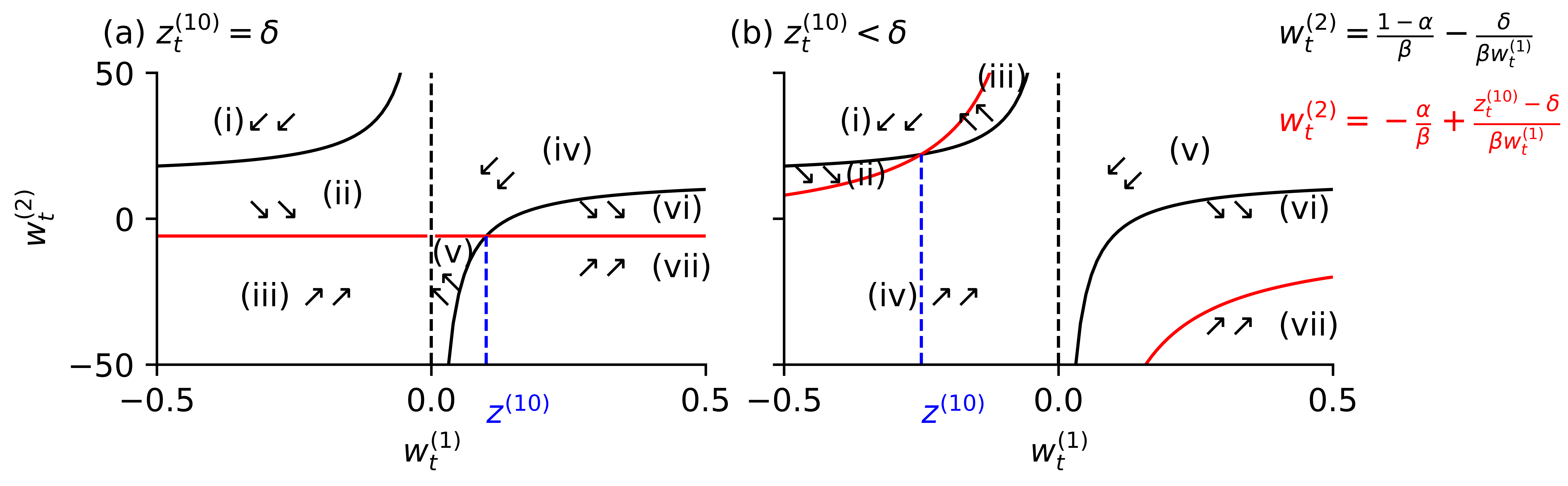}
    \caption{
        Stability diagrams. 
        The solid black and solid red lines represent $\Delta w_t^{(1)}=0$ and $\Delta w_t^{(2)}=0$, respectively. 
        The signs of $\Delta w_t^{(1)}$ and $\Delta w_t^{(2)}$ in (a) and (b) are shown in TABLEs \ref{tabA:2} and \ref{tabA:3}, respectively. 
        In (a) and (b), the stability diagrams of $w_t^{(1)}$ and $w_t^{(2)}$ when $z_t^{(10)}=\delta$ and $z_t^{(10)}>\delta$, respectively, are shown. 
        Note that the intersection point of the nullclines is $(w_t^{(1)},w_t^{(2)}) = (z_t^{(10)}, (1-\alpha)/\beta-\delta/\beta z_t^{(10)})$. 
    }
    \label{figA:stability}
\end{figure*}

To study the divergence and time-dependence of the NARMA10 model without input, we analyzed its attractor. 
We defined new variables for a time-delay system as $z_{t}^{(s)}\equiv y_{t+1-s}~(s=1,2,\ldots,10)$, 
and the model without input ($\mu=\kappa=0$) was rewritten as follows: 
\begin{eqnarray}
    z_{t+1}^{(1)} &=& \alpha z_{t}^{(1)} + \beta z_{t}^{(1)} \sum_{i=1}^{10} z_{t}^{(i)} + \delta, \label{eqA:z1} \\
    z_{t+1}^{(s)} &=& z_{t}^{(s-1)} ~ (s=2,3,\ldots,10). \label{eqA:z2}
\end{eqnarray}
Eq. (\ref{eqA:z1}) makes use of $z_{t}^{(1)}$ and $\sum_{i=1}^{10}z_{t}^{(i)}$; thus, we defined $w_{t}^{(1)} = z_{t}^{(1)}$ and $w_{t}^{(2)} = \sum_{i=1}^{10}z_{t}^{(i)}$, and the discrete derivatives 
$\Delta w_{t}^{(1)} (= w_{t+1}^{(1)} - w_{t}^{(1)})$ and $\Delta w_{t}^{(2)} (= w_{t+1}^{(2)} - w_{t}^{(2)})$ were derived from Eqs. (\ref{eqA:z1}) and (\ref{eqA:z2}) as follows: 
\begin{eqnarray}
    \Delta w_{t}^{(1)} &=& (\alpha-1)w_{t}^{(1)} + \beta w_{t}^{(1)} w_{t}^{(2)} + \delta, \label{eqA:dw1} \\
    \Delta w_{t}^{(2)} &=& \alpha w_{t}^{(1)} + \beta w_{t}^{(1)} w_{t}^{(2)} + \delta - z_{t}^{(10)}. \label{eqA:dw2}
\end{eqnarray}
From Eqs. (\ref{eqA:dw1}) and (\ref{eqA:dw2}), the nullclines can be obtained: 
\begin{eqnarray}
    w_t^{(2)} &=& \frac{1-\alpha}{\beta} - \frac{\delta}{\beta w_t^{(1)}}, \label{eqA:nullcline1} \\
    w_t^{(2)} &=& -\frac{\alpha}{\beta} + \frac{z^{(10)}_{t} - \delta}{\beta w_t^{(1)}}. \label{eqA:nullcline2}
\end{eqnarray}
Note that the intersection point of the nullclines
is a saddle point $(w_t^{(1)}, w_t^{(2)}) = \bigl(z_t^{(10)}, (1-\alpha)/\beta - \delta/\beta z_t^{(10)} \bigr)$. 
Furthermore, we plotted the nullclines on the $w_t^{(1)}$--$w_t^{(2)}$ plane to examine the increase and decrease in $w_t^{(1)}$ and $w_t^{(2)}$ (FIG. \ref{figA:narma10_property}[b] and TABLE \ref{tabA:1}). 
Since $z^{(10)}_{t}$ depends on the time step $t$, the nullcline changes over time 
and three regimes can be distinguished: $z^{(10)}_{t}<\delta$, $z^{(10)}_{t}=\delta$, and $z^{(10)}_{t}>\delta$ (FIGs. \ref{figA:narma10_property}[b] and \ref{figA:stability}). 
The signs of $\Delta w^{(1)}$ and $\Delta w^{(2)}$ are shown in TABLEs \ref{tabA:1}--\ref{tabA:3}. 
From the above results, we found that the model with no input has a fixed point attractor at the saddle point.

\begin{table*}[tb]
    \begin{center}
        \caption{
            The derivative table of $\Delta w_t^{(1)}$ and $\Delta w_t^{(2)}$ in the stability diagram ($z_t^{(10)}>\delta$; FIG. \ref{figA:narma10_property}[b])
        }
        \begin{tabular}{c|c|c||c|c|c||c|c|c||c|c|c} \hline
            \# area & $\Delta w_t^{(1)}$ & $\Delta w_t^{(2)}$ & \# area & $\Delta w_t^{(1)}$ & $\Delta w_t^{(2)}$ & \# area & $\Delta w_t^{(1)}$ & $\Delta w_t^{(2)}$ & \# area & $\Delta w_t^{(1)}$ & $\Delta w_t^{(2)}$ \\ \hline \hline
            (i) & $-$ & $-$ & (iii) & + & + & (v) & $-$ & + & (vii) & + & + \\
            (ii) & + & $-$ & (iv) & $-$ & $-$ & (vi) & + & $-$ &&&\\ \hline
        \end{tabular}
        \label{tabA:1}
    \end{center}
\end{table*}

\begin{table*}[tb]
    \begin{center}
        \caption{
            The derivative table of $\Delta w_t^{(1)}$ and $\Delta w_t^{(2)}$ in the stability diagram ($z_t^{(10)}=\delta$; FIG. \ref{figA:stability}[a])
        }
        \begin{tabular}{c|c|c||c|c|c||c|c|c||c|c|c} \hline
            \# area & $\Delta w_t^{(1)}$ & $\Delta w_t^{(2)}$ & \# area & $\Delta w_t^{(1)}$ & $\Delta w_t^{(2)}$ & \# area & $\Delta w_t^{(1)}$ & $\Delta w_t^{(2)}$ & \# area & $\Delta w_t^{(1)}$ & $\Delta w_t^{(2)}$ \\ \hline \hline
            (i) & $-$ & $-$ & (iii) & + & + & (v) & $-$ & + & (vii) & + & + \\
            (ii) & + & $-$ & (iv) & $-$ & $-$ & (vi) & + & $-$ &&& \\ \hline
        \end{tabular}
        \label{tabA:2}
    \end{center}
\end{table*}

\begin{table*}[tb]
    \begin{center}
        \caption{
            The derivative table of $\Delta w_t^{(1)}$ and $\Delta w_t^{(2)}$ in the stability diagram ($z_t^{(10)}<\delta$; FIG. \ref{figA:stability}[b]) 
        }
        \begin{tabular}{c|c|c||c|c|c||c|c|c||c|c|c} \hline
            \# area & $\Delta w_t^{(1)}$ & $\Delta w_t^{(2)}$ & \# area & $\Delta w_t^{(1)}$ & $\Delta w_t^{(2)}$ & \# area & $\Delta w_t^{(1)}$ & $\Delta w_t^{(2)}$ & \# area & $\Delta w_t^{(1)}$ & $\Delta w_t^{(2)}$ \\ \hline \hline
            (i) & $-$ & $-$ & (iii) & $-$ & + & (v) & $-$ & $-$ & (vii) & + & +\\
            (ii) & + & $-$ & (iv) & + & + & (vi) & + & $-$ & & &\\ \hline
        \end{tabular}
        \label{tabA:3}
    \end{center}
\end{table*}

\subsection{Divergence}
Next, we examined the conditions under which the NARMA10 model diverges. 
Since the nullclines are time-varying due to $z_t^{(10)}$, the increase or decrease of $w_t^{(1)}$ and $w_t^{(2)}$ in each area indicated by FIG. \ref{figA:narma10_property}(b) and TABLE \ref{tabA:1} can change. 
Thus, in the case of no input ($\mu=\kappa=0$), we examined the time step at which $y_t$ diverges with the initial values of $w_0^{(1)}$ and $w_0^{(2)}$. 
The initial values of $y_t$ were set as $y_0=w_0^{(1)}$ and $y_1=y_2=\cdots=y_9=(w_0^{(2)}-w_0^{(1)})/9$. 
As shown in FIG. \ref{figA:narma10_property}(c), the combination of initial values $(w_0^{(1)},w_0^{(2)})$ where $y_t$ does not diverge is distributed in a complex manner. 
We attribute the difference between the theoretical and numerical basins of attraction to the change in $z_t^{(10)}$, which produced a complicated distribution of the nullclines (see FIGs. \ref{figA:narma10_property}[b] and \ref{figA:stability}, and TABLEs \ref{tabA:1}--\ref{tabA:3}).

In the presence of input, $y_t$ diverges due to the initial condition. 
As shown in FIG. \ref{figA:narma10_property}(d), the model with input $u_t\in[0,0.4]$ ($\mu=\kappa=0.2$) diverges at an initial value similar to the one in the case of no input. 
FIG. \ref{figA:narma10_property}(e) shows the bifurcation diagram of $y_t$ given an initial value at which $y_t$ converges to the fixed-point. 
Since the fixed-point around which $y_t$ fluctuates is the saddle point, $y_t$ can diverge with the given input. 
To investigate the divergence conditions caused by the input, we ran the model over $10^6$ time steps and examined the time step at which $y_t$ diverges by altering the initial values $y_0=\cdots=y_9=\psi$ and input intensity $\sigma$. 
FIG. \ref{figA:narma10_property}(f) shows that $y_t$ diverges to infinity when $\psi$ or $\sigma$ exceeds three thresholds: (i) $\psi<-5.15$, (ii) $1.239<\psi$, and (iii) $\sigma>0.45$. 
In the (i) and (ii) cases, $y_t$ diverges at a shorter time step ($t<100$) than in (iii) because the divergence is caused by the initial value $\psi$. 
However, in case of (iii), $y_t$ diverges according to the input. 
After converging to the fixed-point, $y_t$ can diverge successively when receiving large positive inputs. 
According to the time steps in FIG. \ref{figA:narma10_property}(f), when $y_t$ diverges as the runtime becomes longer, $y_t$ diverges with smaller $\sigma$. 
Consequently, the threshold (iii) is the boundary that depends on the input time-series. 
These results suggest that $y_t$ in the vicinity of the fixed-point can diverge depending on the input.

In FIG. \ref{fig:ipc_narma10}, $p$ expresses the probability of not diverging as a function of $\sigma$ for different random series $\{\zeta_t\}$. 
As previously shown, once $y_t$ converges to the fixed-point, the model stochastically diverges due to the input, 
and the probability $p$ depends on $\sigma$; 
thus, even though $p$ is high, $y_t$ can potentially diverge. 
For example, two typical ranges of input, $u_t\in[0,0.5]$ and $[0,1]$, have been used for the NARMA10 task; 
however, $y_t$ diverges in both cases (FIG. \ref{fig:ipc_narma10}[b], $\sigma=0.5,1.0$) 
because $\sigma$ is relevant to the average time until divergence, and the runtime, $10^6$ time steps, is much longer than the time used for the benchmark task. 
Consequently, the divergence probability of $y_t$, $p$, changes depending on the parameter $\sigma$. 

Therefore, although the NARMA10 model produces the fixed-point attractor, it can potentially diverge depending on the initial values, input time-series, and parameter settings.

\subsection{Time-variance analysis}
Finally, we investigated the time-dependence of the NARMA10 model.
The state of a dynamical system receiving noise input can transit from non-chaos to chaos, an effect referred to as noise-induced chaos \cite{crutchfield1982fluctuations}. 
As chaotic behavior is exhibited by a time-variant system, we investigated whether the system is chaotic or ordered by calculating the maximum Lyapunov exponent $\lambda_1$.
Thus, we derived the Lyapunov spectrum of the model $\lambda_i~(i=1,2,\ldots,10)$ based on the 10-dimensional time-delay system in Eqs. (\ref{eqA:z1}) and (\ref{eqA:z2}). 
We expressed $z_{t}^{(s)}~(s=1,\ldots,10)$ as a vector $\bm{z}_t = [z^{(1)}_t\cdots z^{(10)}_{t}]^\top$, 
and the Jacobian matrix of Eq. (\ref{eq:narma10}) ${\bm J}_t\in\mathbb{R}^{10\times 10}$ with respect to ${\bm z}_t$ can be written as follows: 
\begin{eqnarray}
    {\bm J}_t = \frac{\partial {\bm z}_{t+1}}{\partial {\bm z}_{t}} = 
    \begin{bmatrix}
        X_t & Y_t & \cdots & Y_t & Y_t \\
        1 &   &   &   & \\
          & 1 &   &   & \\
          &   & \ddots & & \\
          &   &   & 1 & \\
    \end{bmatrix}, \label{eqA:dzdz}
\end{eqnarray}
where $X_t=\alpha+2\beta z_{t}^{(1)}+\beta\sum_{i=2}^{10} z_{t}^{(i)}$ and $Y_t = \beta z_{t}^{(1)}$. 
Using the Jacobian matrices, the Lyapunov spectrum was computed as follows: 
\begin{eqnarray}
    \lambda_{i} = \frac{1}{T} \sum_{t=1}^{T} \ln{ \rho_i({\bm J}_{t+M-1}{\bm J}_{t+M-2}\cdots{\bm J}_t) }~(i=1,2,\ldots,10),\nonumber\\
    \label{eqA:lyapunov}
\end{eqnarray}
where $\rho_i({\bm J}_{t+M-1}{\bm J}_{t+M-2}\cdots{\bm J}_t)$ is the $i^{\rm th}$ singular value of matrix ${\bm J}_{t+M-1}{\bm J}_{t+M-2}\cdots{\bm J}_t$, while $T$ and $M$ were set to 6000 and 40, respectively. 
FIG. \ref{figA:narma10_property}(g) shows the three largest Lyapunov spectra, $\lambda_1$, $\lambda_2$, and $\lambda_3$, all of which are negative relative to $\sigma$, indicating that the system does not demonstrate chaos. 
Therefore, the NARMA10 model is not a chaotic and time-variant system.

The divergence and time-invariance analysis results revealed that the NARMA10 model converges to the fixed-point and varies in the vicinity of the point. 
We considered the $y_t$ fluctuating around the fixed-point to be time-invariant.

\renewcommand{\theequation}{E\arabic{equation}}
\setcounter{equation}{0}
\renewcommand{\thetable}{E\arabic{table}}
\setcounter{table}{0}
\renewcommand{\thefigure}{E\arabic{figure}}
\setcounter{figure}{0}

\section{Interconvertibility of the PC expansion and IPC}
To clearly demonstrate that the PC expansion and IPC are interconvertible, we derived an approximate model that has nearly the same breakdown of the IPC as the original breakdown using Legendre-chaoses. 
From the above capacity analysis, we narrowed the polynomial terms to $P_1(\zeta_{t-s})~(s=1,2,\ldots)$ and $P_1(\zeta_{t-s})P_1(\zeta_{t-s-9})~(s=1,2,\ldots)$, which yielded significantly greater capacities. 
The expanded state is expressed as follows: 
\begin{eqnarray}
    y_{t} = p + \sum_{s\in\mathcal{N}_1} q_{s} P_{1}(\zeta_{t-s}) + \sum_{s\in\mathcal{N}_2} r_{s} P_1 (\zeta_{t-s}) P_1 (\zeta_{t-s-9}),\nonumber\\ 
\end{eqnarray}
where $p$, $q_{s}$, and $r_{s}$ are coefficients for the Legendre-chaoses $P_0=1$, $P_1(\zeta_{t-s})$, and $P_1(\zeta_{t-s})P_1(\zeta_{t-s-9})$, respectively, 
while $P_1(\zeta)=\zeta$. 
$\zeta_t$ follows a uniform distribution in $[-1,1]$, 
and $\mathcal{N}_1$ and $\mathcal{N}_2$ represent the sets of delayed time steps $s$ for $P_1(\zeta_{t-s})$ and $P_1(\zeta_{t-s})P_1(\zeta_{t-s-9})$, respectively. 
Let the normalized Legendre-chaoses be $\phi_0=\frac{1}{\sqrt{T}}$, $\phi_{1,t}^{(s)}=\frac{P_1(\zeta_{t-s})}{\sqrt{\sum_{t=1}^TP(\zeta_{t-s})^2}}$, and $\phi_{2,t}^{(s)}=\frac{P_1(\zeta_{t-s})P_1(\zeta_{t-s-9})}{\sqrt{\sum_{t=1}^T\{P(\zeta_{t-s})P(\zeta_{t-s-9})\}^2}}$, and the state is represented as follows:
\begin{eqnarray}
    y_{t} = \hat{p}\phi_0 + \sum_{s\in\mathcal{N}_1} \hat{q}_{s} \phi_{1,t}^{(s)} + \sum_{s\in\mathcal{N}_2} \hat{r}_{s} \phi_{2,t}^{(s)}, \label{eqA:novel_model}
\end{eqnarray}
where $\hat{p}=p\sqrt{T}$, $\hat{q}_s=q_s\sqrt{\sum_{t=1}^TP(\zeta_{t-s})^2}$, and $\hat{r}_s=r_s\sqrt{\sum_{t=1}^T\{P(\zeta_{t-s})P(\zeta_{t-s-9})\}^2}$ are the modified coefficients for $p$, $q_s$, and $r_s$, respectively.

\begin{figure*}[tb]
    \centering
    \includegraphics[scale=0.85]{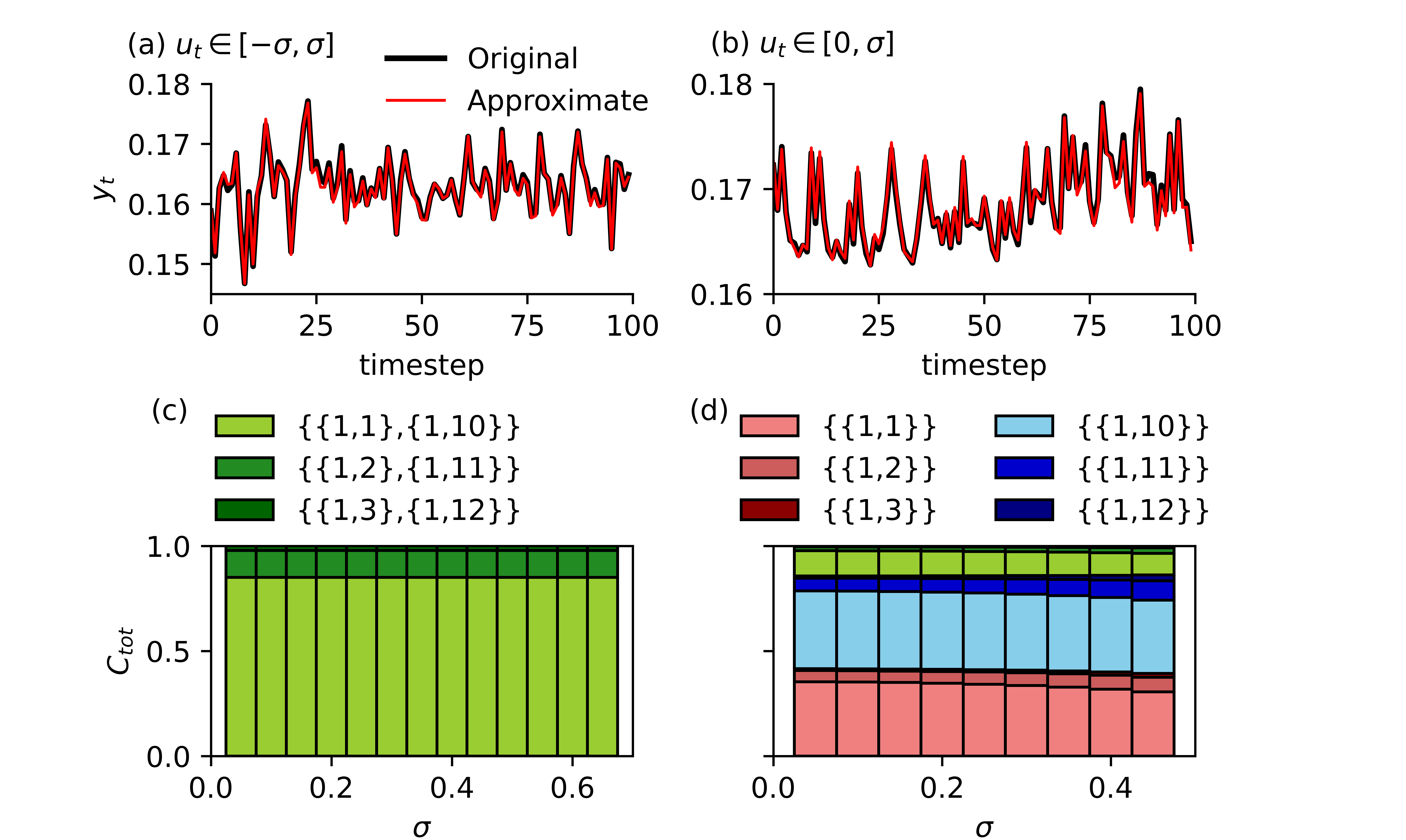}
    \caption{
        The proposed model for the benchmark task. 
        Frames (a) and (b) show the time-series data with $u_t\in [-\sigma,\sigma]$ and $[0,\sigma]$, respectively. 
        The solid black and solid red lines represent the output of the NARMA10 and our models, respectively. 
        Panels (c) and (d) show the IPC with $u_t\in[-\sigma,\sigma]$ and $[0,\sigma]$, respectively.  
        The labels represent combinations of $\{\{n_s,s\}\}$, where $n_s$ is the degree of the polynomial, and $s$ is the delayed time step of the input. Here, the desired output is $\prod_s P_{n_s}(\zeta_{t-s})$.
    }
    \label{figA:approximate_narma10}
\end{figure*}

Detrending the state and using Eq. (\ref{eq:IPC1}), the IPCs for $\phi_{1,t}^{(s)}$ and $\phi_{2,t}^{(s)}$ become
\begin{eqnarray}
    C = \frac{q_s^2}{\sum_{s\in\mathcal{N}_1}q_s^2 + \sum_{s\in\mathcal{N}_2}r_s^2} \label{eqA:IPC_q}
\end{eqnarray}
and 
\begin{eqnarray}
    C = \frac{r_s^2}{\sum_{s\in\mathcal{N}_1}q_s^2 + \sum_{s\in\mathcal{N}_2}r_s^2}, \label{eqA:IPC_r}
\end{eqnarray}
respectively. 
Therefore, Eqs. (\ref{eqA:IPC_q}) and (\ref{eqA:IPC_r}) show that each IPC is the normalized squared coefficient in the polynomial chaos expansion. 
To demonstrate this model, we employ $\mathcal{N}_1=\{1,2,3,10,11,12\}$ and $\mathcal{N}_2=\{1,2,3\}$. 
As shown in FIG. \ref{figA:approximate_narma10}(a) and (b), for $u_t \in [-\sigma,\sigma]$ and $[0,\sigma]$, respectively, the approximate model successfully reproduced the original NARMA10 model. 
Furthermore, as in FIG. \ref{figA:approximate_narma10}(c) and (d), we confirmed that the IPC breakdown of the approximate model reproduced the original breakdown.

\subsection{Derivation}
Here, we derive the following equations: 
\begin{eqnarray}
    y_{t} &=& p + \sum_{s\in\mathcal{N}_1} q_{s} P_{1}(\zeta_{t-s}) \nonumber \\
    & & \ \ + \sum_{s\in\mathcal{N}_2} r_{s} P_1 (\zeta_{t-s}) P_1 (\zeta_{t-s-9}), \label{eqA:novel_model} \\
    p &=& \frac{1-\alpha}{20\beta} - \sqrt{ \left( \frac{1-\alpha}{20\beta} \right)^2 - \frac{\gamma \mu^2 + \delta}{10\beta} }, \label{eqA:p}
\end{eqnarray}
\begin{eqnarray}
    q_{s} &=& 
    \begin{cases}
        \gamma \mu \kappa ~ (s=1) \\
        (\alpha+10\beta p)q_{s-1} + \sum_{j=0}^{s-2}\beta p q_{s-j-1} \\ (s=2,3,\ldots,9) \\
        \gamma \mu \kappa + (\alpha+10\beta p)q_{s-1} + \sum_{j=0}^{s-2}\beta p q_{s-j-1} \\
        (s=10) \\
        (\alpha+10\beta p)q_{s-1} + \sum_{j=0}^{9}\beta p q_{s-j-1} \\
        (s=11,12,\ldots)
    \end{cases}, \nonumber\\
    \label{eqA:q}
\end{eqnarray}
\begin{eqnarray}
    r_{s} &=& 
    \begin{cases}
        \gamma \kappa^2 ~ (s=1) \\
        (\alpha+10\beta p)r_{s-1} + \beta q_{s-1}\sum_{j=0}^9 q_{s+8-j} \\
        + \beta\sum_{j=0}^{s-2} (pr_{s-j-1}+q_{s+8}q_{s-j-1}) \\ (s=2,3,\ldots,10)\\
        (\alpha+10\beta p)r_{s-1} + \beta q_{s-1}\sum_{j=0}^9 q_{s+8-j} \\
        + \beta\sum_{j=0}^{9} (pr_{s-j-1}+q_{s+8}q_{s-j-1}) \\ (s=11,12,\ldots)
    \end{cases}, \nonumber\\
    \label{eqA:r}
\end{eqnarray}
where $\mathcal{N}_1$ and $\mathcal{N}_2$ represent the sets of delayed time steps $s\in\mathbb{N}$ for $P_1(\zeta_{t-s})$ and $P_1(\zeta_{t-s})P_1(\zeta_{t-s-9})$, respectively.

First, $y_t$ in Eq. (\ref{eq:narma10}) is expanded using the Legendre-chaoses of input time-series $\zeta_{t-s}~(s=1,2,\ldots,t)$ with time-varying coefficients as follows: 
\begin{eqnarray}
    y_{t} &=& p_t + \sum_{s=1}^{t} q_{t,s} P_{1}(\zeta_{t-s}) \nonumber \\
    &+& \sum_{s=1}^{t} r_{t,s} P_1 (\zeta_{t-s}) P_1 (\zeta_{t-s-9}) + \cdots, \label{eqA:xt}
\end{eqnarray}
where $p_t$ denotes a time-varying term independent of $\zeta_{t-s}~(s=1,2,\ldots,t)$, 
and $q_{t,s}$ and $r_{t,s}$ are the $s^{\rm th}$ coefficients of $P_{1}(\zeta_{t-s})$ and $P_1 (\zeta_{t-s}) P_1 (\zeta_{t-s-9})$, respectively. 
The NARMA10 model with $\zeta_t$ can be expressed as 
\begin{eqnarray}
    y_{t+1} &=& \alpha y_t + \beta y_t \sum_{j=0}^{9}y_{t-j} \nonumber \\
    &+& \gamma (\mu + \sigma \zeta_t) (\mu + \sigma \zeta_{t-9}) + \delta. \label{eqA:narma10}
\end{eqnarray}
According to Eqs. (\ref{eqA:xt}) and (\ref{eqA:narma10}), $y_{t+1}$ is rewritten as 
\begin{eqnarray}
    y_{t+1} &=& 
    \left( \alpha p_t + \beta p_t \sum_{j=0}^{9} p_{t-j} + \gamma \mu^2 + \delta \right) \nonumber \\
    &+& \gamma \mu \kappa \left( P_1 (\zeta_{t}) + P_1 (\zeta_{t-9}) \right) \nonumber \\
    &+& \alpha \sum_{s=1}^{t} q_{t,s} P_1 (\zeta_{t-s}) \nonumber \\
    &+& \beta p_t \sum_{j=0}^9 \sum_{s=1}^{t-j} q_{t-j,s} P_1 (\zeta_{t-s-j}) \nonumber \\
    &+& \beta \sum_{j=0}^9 p_{t-j} \sum_{s=1}^{t} q_{t,s} P_1 (\zeta_{t-s}) \nonumber \\
    &+& \gamma\kappa^2 \zeta_t \zeta_{t-9} \nonumber\\
    &+& \alpha \sum_{s=1}^{t} r_{t,s} P_1 (\zeta_{t-s}) P_1 (\zeta_{t-s-9}) \nonumber\\
    &+& \beta p_t \sum_{j=0}^9 \sum_{s=1}^{t-j} r_{t-j,s} P_1 (\zeta_{t-s-j}) P_1 (\zeta_{t-s-j-9}) \nonumber\\
    &+& \beta \sum_{j=0}^9 p_{t-j} \sum_{s=1}^{t} r_{t,s} P_1 (\zeta_{t-s}) P_1 (\zeta_{t-s-9}) \nonumber\\
    &+& \sum_{s=1}^{t} q_{t,s} P_1 (\zeta_{t-s}) P_1 (\zeta_{t-s}) \sum_{j=0}^9 \sum_{s=1}^{t-j} q_{t-j,s} P_1 (\zeta_{t-s-j}) \nonumber \\
    &+& \cdots, \label{eqA:xt+1_1}
\end{eqnarray}
where $P_1 (\zeta) = \zeta$. 
When increasing $t$ by one in Eq. (\ref{eqA:xt}), the following equation is obtained: 
\begin{eqnarray}
    y_{t+1} &=& p_{t+1} + \sum_{s=1}^{t+1} q_{t+1,s} P_{1}(\zeta_{t+1-s}) \nonumber \\
    &+& \sum_{s=1}^{t+1} r_{t+1,s} P_1 (\zeta_{t+1-s}) P_1(\zeta_{t+1-s-9}) + \cdots. \nonumber\\
    \label{eqA:xt+1_2}
\end{eqnarray}
Equating the coefficients in Eqs. (\ref{eqA:xt+1_1}) and (\ref{eqA:xt+1_2}) yields: 
\begin{eqnarray}
    p_{t+1} &=& \alpha p_t + \sum_{j=0}^9 \beta p_t p_{t-j} + \gamma \mu^2 + \delta, \label{eqA:pt+1}
\end{eqnarray}
\begin{eqnarray}
    q_{t+1,s} &=& 
    \begin{cases}
        \gamma \mu \kappa ~ (s=1) \\
        \left( \alpha + \beta \sum_{j=0}^{9}p_{t-j} \right)q_{t,s-1} \\
        + \sum_{j=0}^{s-2} \beta p_t q_{t-j,s-j-1} ~ (s=2,3,\ldots,9) \\
        \gamma \mu \kappa + \left( \alpha + \beta \sum_{j=0}^{9}p_{t-j} \right)q_{t,s-1} \\
        + \sum_{j=0}^{s-2} \beta p_t q_{t-j,s-j-1} ~ (s=10) \\
        \left( \alpha + \beta \sum_{j=0}^{9}p_{t-j} \right)q_{t,s-1} \\
        + \sum_{j=0}^{9} \beta p_t q_{t-j,s-j-1} ~ (s=11,12,\ldots)
    \end{cases}, \nonumber\\ \label{eqA:qt+1} 
\end{eqnarray}
\begin{eqnarray}
    r_{t+1,s} &=& 
    \begin{cases}
        \gamma \kappa^2 ~ (s=1) \\
        \left( \alpha + \beta \sum_{j=0}^9 p_{t-j} \right)r_{t,s-1} \\
        + \beta q_{t,s-1}\sum_{j=0}^9 q_{t-j,s+8-j} \\
        + \beta \sum_{j=0}^{s-2} (p_t r_{t-j,s-j-1}+q_{t,s+8}q_{t-j,s-j-1}) \\ (s=2,3,\ldots,10)\\
        \left( \alpha+\beta\sum_{j=0}^9 p_{t-j} \right)r_{t,s-1} \\
        + \beta q_{t,s-1}\sum_{j=0}^9 q_{t-j,s+8-j} \\
        + \beta\sum_{j=0}^{9} (p_t r_{t-j,s-j-1}+q_{t,s+8}q_{t-j,s-j-1}) \\ (s=11,12,\ldots)
    \end{cases}. \nonumber\\ \label{eqA:rt+1}
\end{eqnarray}
According to Eq. (\ref{eqA:pt+1}), $p_t$ has a stable and an unstable equilibrium point. 
If $p_t<(1-\alpha)/20\beta + \sqrt{((1-\alpha)/20\beta)^2-(\gamma \mu^2 + \delta)/10\beta}$, it converges to the stable point. 
When $t$ is large enough, $p_t$ converges to 
\begin{eqnarray}
    p=\lim_{t \rightarrow \infty} p_{t} = \frac{1-\alpha}{20\beta} - \sqrt{ \left( \frac{1-\alpha}{20\beta} \right)^2-\frac{\gamma \mu^2 + \delta}{10\beta}}. \nonumber\\ \label{eqA:p}
\end{eqnarray}
According to Eq. (\ref{eqA:qt+1}), $q_{t,s}$ also converges to
\begin{eqnarray}
    q_{s} &=& \lim_{t \rightarrow \infty} q_{t+1,s} \nonumber \\
    &=& 
    \begin{cases}
        \gamma \mu \kappa ~ (s=1) \\
        (\alpha+10\beta p)q_{s-1} + \sum_{j=0}^{s-2}\beta p q_{s-j-1} \\ (s=2,3,\ldots,9) \\
        \gamma \mu \kappa + (\alpha+10\beta p)q_{s-1} + \sum_{j=0}^{s-2}\beta p q_{s-j-1} \\
        (s=10) \\
        (\alpha+10\beta p)q_{s-1} + \sum_{j=0}^{9}\beta p q_{s-j-1} \\
        (s=11,12,\ldots)
    \end{cases}. 
    \label{eqA:qi} \nonumber\\ 
\end{eqnarray}
In the same manner, $r_{t,s}$ converges to
\begin{eqnarray}
    r_{s} &=& \lim_{t \rightarrow \infty} r_{t+1,s} \nonumber \\
    &=&
    \begin{cases}
        \gamma \kappa^2 ~ (s=1) \\
        (\alpha+10\beta p)r_{s-1} + \beta q_{s-1}\sum_{j=0}^9 q_{s+8-j} \\
        + \beta\sum_{j=0}^{s-2} (pr_{s-j-1}+q_{s+8}q_{s-j-1}) \\ (s=2,3,\ldots,10)\\
        (\alpha+10\beta p)r_{s-1} + \beta q_{s-1}\sum_{j=0}^9 q_{s+8-j} \\
        + \beta\sum_{j=0}^{9} (pr_{s-j-1}+q_{s+8}q_{s-j-1}) \\ (s=11,12,\ldots)
    \end{cases}. 
    \label{eqA:ri} \nonumber\\ 
\end{eqnarray}
Therefore, when $t$ is large enough, and Eq. (\ref{eqA:xt}) is approximated with the constant term $p$ and the Legendre-chaoses of $P_1(\zeta_{t-s})$ and $P_1(\zeta_{t-s})P_1(\zeta_{t-s-9})$, whose delayed time steps $s$ are limited to sets $\mathcal{N}_1$ and $\mathcal{N}_2$, respectively, Eqs. (\ref{eqA:novel_model})--(\ref{eqA:r}) are obtained.

\renewcommand{\theequation}{F\arabic{equation}}
\setcounter{equation}{0}
\renewcommand{\thetable}{F\arabic{table}}
\setcounter{table}{0}
\renewcommand{\thefigure}{F\arabic{figure}}
\setcounter{figure}{0}

\section{\label{sec:culture}How to compose the dissociated culture reservoir}
All experiments were approved by the ethical committee of the University of Tokyo and followed the ``Guiding Principles for the Care and Use of Animals in the Field of Physiological Science'' estalished by the Physiological Society of Japan. 
Embryonic rat cortices were dissected from E18 rats and used for cortical cell cultures. 
The cortices were dissociated in 2 mL of 0.25\% trypsin-ethylenediaminetetraacetic acid (Trypsin-EDTA, Life Technologies), from which cells were isolated by trituration, and 38,000 cells were seeded on each microelectrode array (MEA; MaxWell Biosystems). 
For cell adhesion, 5 mL of 0.05\% Polyethileneimine (PEI; Sigma-Aldrich) and 5 ${\rm \mu}$l of 0.02 mg/ml Laminin (Sigma-Aldrich) were used before plating the cells. 
Then, after 24 hours, the plating media\cite{brewer1993optimized} were changed to growth media\cite{potter2001new}. 
The plating media were composed of Neurobasal $850\ {\rm \mu l}$ (Life Technologies), 10\% horse serum (HyClone), 0.5 mM GlutaMAX (Life Technologies), and 2\% B27 (Life Technologies). 
The growth media were composed of DMEM $850 {\rm \mu l}$ (Life Technologies), 10\%horse serum (HyClone), 0.5 mM GlutaMAX (Life Technologies), and 1 mM sodium pyruvate (Life Technologies). 
All experiments were conducted in an incubator at 37$^\circ$C and 5\% CO$_2$. 
The MEAs were sealed with a lid to prevent water evaporation and invasion of bacteria and fungus.

The MEA had 26,400 electrodes, which were placed 17.5 $\rm\mu$m apart and arranged in a 120 $\times$ 220 grid. 
The MEA can simultaneously use up to 1,024 of 26,400 electrodes. 
We selected electrodes with a high firing rate as measurement electrodes and electrodes on which the axon places, as stimulation electrodes. 
We applied bipolar pulse stimuli with an amplitude of $\zeta_t$, which followed a normal distribution with mean $\mu$ and standard deviation $\sigma$, and of an interpulse interval (IPI) of 10, 20, and 30 ms to the stimulation electrodes. 
Furthermore, a 6$^{\rm th}$-order Butterworth bandpass filter and zero-phase IIR filter were applied to the voltage traces observed from the measurement electrodes to extract 300--3000 Hz components. 
At all electrodes, stimulus-induced artifacts were removed by eliminating traces $\pm$2 ms from the stimulus times. 
The standard deviation of extracted signals was calculated as follows \cite{quiroga2004unsupervised}: 
\begin{eqnarray}
    \sigma = {\rm median} \biggl\{ \frac{|{\bm x}|}{0.6745} \biggr\}. \label{eq:std}
\end{eqnarray}
If the amplitude of an extracted signal exceeded $4\sigma$, the value of the spike train was set to one; otherwise, it was set to zero. 
As the measurement frequency was 20 kHz, the above spike train was separated by a 1-ms time bin, and if one or more spikes appeared in one bin, the modified spike train was set to one; otherwise it was set to zero. 
The train was divided into bins by IPI-width, and the number of spikes in the bin was used for the state ${\bm x}_t$.

\bibliography{main}

\end{document}